\documentclass[10pt,twocolumn,letterpaper]{article}

\usepackage[pagenumbers]{cvpr} 

\newcommand{\red}[1]{{\color{red}#1}}

\usepackage{makecell}
\usepackage{color}
\usepackage{multirow}
\usepackage{amsmath}
\usepackage{enumitem}
\usepackage{graphicx}
\usepackage{caption}

\newcommand{\blue}[1]{\textcolor{blue}{#1}} 
\usepackage[ruled,vlined,linesnumbered]{algorithm2e}
\definecolor{mygray1}{gray}{.95}
\definecolor{mygray2}{gray}{.9}
\definecolor{mygray3}{gray}{.95}
\definecolor{mygray}{gray}{.9}
\usepackage{colortbl}

\usepackage{marvosym}

\definecolor{cvprblue}{rgb}{0.21,0.49,0.74}
\usepackage[pagebackref,breaklinks,colorlinks,allcolors=cvprblue]{hyperref}

\def\paperID{****} 
\def\confName{CVPR}
\def\confYear{2025}

\title{
\includegraphics[height=1.3em]{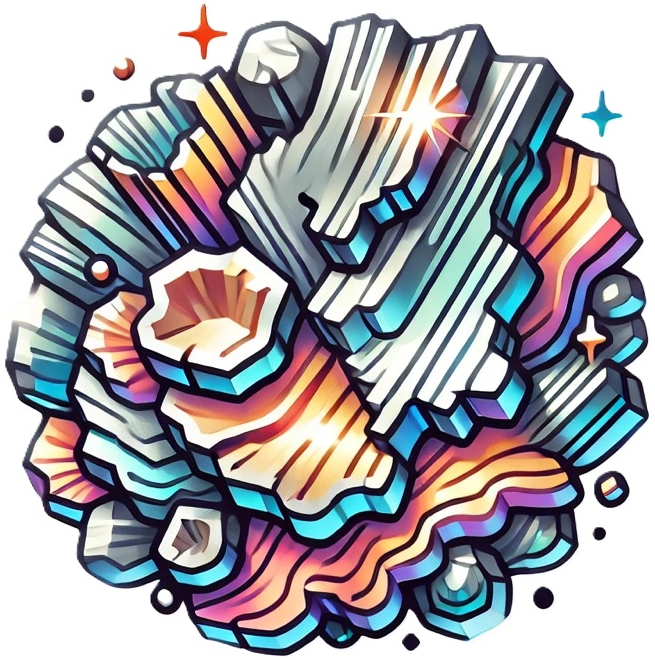} 
MICAS: Multi-grained In-Context Adaptive Sampling for \\ 3D Point Cloud Processing}

\author{
  Feifei Shao$^{1}$\thanks{Feifei Shao and Ping Liu are co-first authors with equal contributions.}, Ping Liu$^{2 } $\footnotemark[1], Zhao Wang$^3$, Yawei Luo$^4$\thanks{Yawei Luo is the corresponding author.}, Hongwei Wang$^1$, Jun Xiao$^5$ \\
  \small $^1$ Zhejiang University-University of Illinois Urbana-Champaign Institute, Zhejiang University, China \; \\
  \small $^2$ Computer Science and Engineering, University of Nevada, Reno, USA \; \\
  \small $^3$ Ningbo Innovation Center, Zhejiang University, China \; 
  \small $^4$ School of Software Technology, Zhejiang University, China \; \\
  \small $^5$ College of Computer Science and Technology, Zhejiang University, China \; \\
  \small \texttt{\{sff, zhao\_wang, yaweiluo\}@zju.edu.cn, pino.pingliu@gmail.com, } \\
  \small \texttt{hongweiwang@intl.zju.edu.cn, junx@cs.zju.edu.cn} \\
  }

\begin{document}
\maketitle

\begin{abstract}
   Point cloud processing (PCP) encompasses tasks like reconstruction, denoising, registration, and segmentation, each often requiring specialized models to address unique task characteristics. While in-context learning (ICL) has shown promise across tasks by using a single model with task-specific demonstration prompts, its application to PCP reveals significant limitations. We identify inter-task and intra-task sensitivity issues in current ICL methods for PCP, which we attribute to inflexible sampling strategies lacking context adaptation at the point and prompt levels. To address these challenges, we propose MICAS, an advanced ICL framework featuring a multi-grained adaptive sampling mechanism tailored for PCP. MICAS introduces two core components: task-adaptive point sampling, which leverages inter-task cues for point-level sampling, and query-specific prompt sampling, which selects optimal prompts per query to mitigate intra-task sensitivity. To our knowledge, this is the first approach to introduce adaptive sampling tailored to the unique requirements of point clouds within an ICL framework. Extensive experiments show that MICAS not only efficiently handles various PCP tasks but also significantly outperforms existing methods. Notably, it achieves a remarkable $4.1\%$ improvement in the part segmentation task and delivers consistent gains across various PCP applications.
\end{abstract}    

\section{Introduction}

\begin{figure*}[t]
  \centering
  \includegraphics[width=1.0\linewidth]{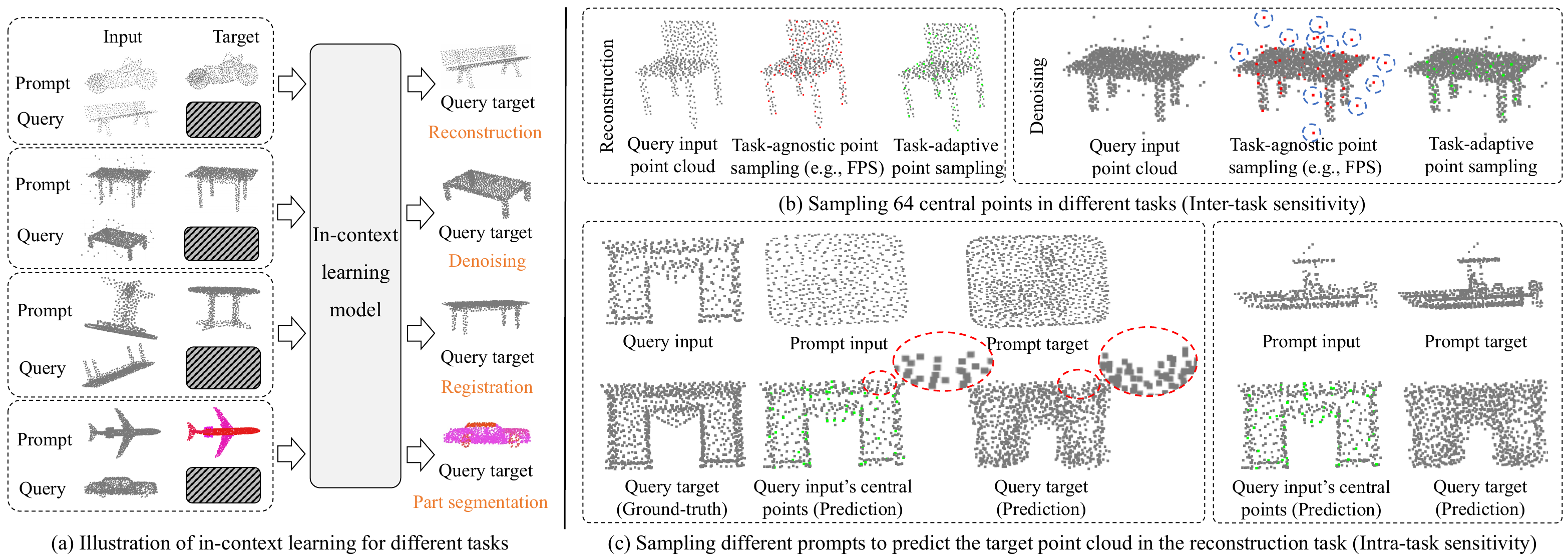}
  \vspace{-1.6em}
  \caption{Inter-task and intra-task sensitivities in in-context learning. The red and green points are sampled using Farthest Point Sampling (FPS) and task-adaptive point sampling, respectively. The blue circles indicate erroneous sampling points, while the red ovals highlight missing points in the predicted point cloud, caused by the absence of a central point within the region. (Zoom in for more details)}
  \label{intro}
  \vspace{-1.0em}
\end{figure*}

Deep learning has greatly advanced 3D point cloud processing, tackling tasks like semantic segmentation~\cite{shao2022active, kolodiazhnyi2024oneformer3d}, registration~\cite{yang2020teaser, zhang2024svc}, reconstruction~\cite{mandikal2019dense, huang2024surface}, and denoising~\cite{luo2021score, wei2024pathnet}. 
However, achieving high performance often requires separate models for each task, increasing complexity and resource demands. 
Multi-task Learning (MTL)~\cite{zhang2021survey, zhao2024robust, shan2023gpa} attempts to reduce this burden by training models to handle multiple tasks simultaneously, but it struggles with performance trade-offs and complex parameter tuning. 
In contrast, In-context Learning (ICL)~\cite{fang2024explore, xu2024context, jiang2025dg} offers a simpler approach, using only a few prompts to guide a single model in performing multiple tasks without changing its parameters~\cite{brown2020language, xu2024context,bar2022visual, wang2023images, luo2024large, wang2023seggpt}. 

Despite these advancements, recent efforts to extend ICL to 3D point cloud processing~\cite{fang2024explore, liu2024point} reveal significant limitations. Specifically, these studies have not fully addressed critical challenges associated with conventional point cloud sampling techniques used in the ICL framework. Taking Figure~\ref{intro} (a) for example, the ICL framework manages multiple point cloud tasks,
each with distinct preferences for point cloud sampling. 
However, conventional sampling methods struggle to adapt equally well to these diverse tasks simultaneously.
These gaps, particularly in adapting sampling techniques to task-specific and prompt-specific contexts, hinder overall performance and reliability. 

To overcome these limitations, our work tackles two critical issues:
\textbf{1) Inter-task Sensitivity:} As illustrated in Figure~\ref{intro} (b), task-agnostic sampling strategies, \eg, Farthest Point Sampling (FPS), may perform differently across \emph{different tasks}, such as reconstruction and denoising. 
This difference arises because FPS tends to prioritize outliers, often leading to the selection of noisy points.
This issue underscores the urgent need for a methodology that effectively integrates task information into the sampling process. \textbf{2) Intra-task Sensitivity:} Depicted in Figure~\ref{intro} (c), variations in prompts for the \emph{same task} can yield divergent sampling outcomes, resulting in inconsistent experimental results.
This highlights the need to replace generic prompts with carefully curated, query-specific prompts.

To effectively address the inter-task and intra-task sensitivity issues, we introduce a novel Multi-grained In-Context Adaptive Sampling mechanism, dubbed \textbf{MICAS}, for 3D point cloud in-context learning. As shown in Figure~\ref{overview} (b), MICAS comprises two integral components: task-adaptive point sampling and query-specific prompt sampling.

For inter-task sensitivity, task-adaptive point sampling enables adaptive sampling by interpreting various prompts, operating in two stages: prompt understanding and Gumbel sampling.
First, the prompt understanding phase extracts essential task features from the prompt and corresponding point features from point clouds, providing a basis for informed sampling. 
However, traditional discrete sampling methods fail to support gradient-based optimization, jeopardizing both the efficiency and effectiveness of the learning process.
To address this, the Gumbel sampling phase leverages the Gumbel-softmax~\cite{jang2016categorical}, transforming discrete sampling into a differentiable operation and enabling a fully learnable and efficient sampling process~\cite{wen2023learnable}.

To mitigate the intra-task sensitivity caused by prompt variability, we integrate a query-specific prompt sampling module. 
This module selects the most effective prompt by ranking the sampling probabilities, which are aligned to the inference performance.
Specifically, we first predict sampling probabilities for each prompt by analyzing 
queries and prompts, followed by aligning these probabilities with the in-context learning model's performance. During inference, the ``best-performing'' prompt is selected based on these probabilities among strategically chosen candidate prompts.

We evaluate our design on a benchmark~\cite{fang2024explore} comprising multiple existing datasets~\cite{chang2015shapenet, yi2016scalable}, covering four distinct point cloud tasks with five levels of difficulty each.
Our comprehensive evaluation demonstrates the efficacy of MICAS in addressing both inter-task and intra-task sensitivity issues. 
In addition, these results highlight the practical advantages of MICAS, including enhanced adaptability to diverse 3D point cloud tasks and improved robustness across various ICL model variants. 
The contributions of this paper are summarized as follows:

\begin{itemize}[leftmargin=15pt]
    \item We propose a novel multi-grained in-context adaptive sampling mechanism, MICAS, that effectively addresses inter-task and intra-task sensitivity issues in 3D point cloud in-context learning.
    \item MICAS integrates two key components: task-adaptive point sampling and query-specific prompt sampling. The former dynamically adjusts to task-specific needs at the point level, while the latter refines prompt selection to minimize intra-task variability. Together, these components enable adaptive and efficient sampling across diverse 3D point cloud tasks.
    \item Extensive experiments demonstrate that MICAS not only simplifies training and efficiently handles multiple tasks but also achieves substantial performance gains over previous state-of-the-art methods, including a notable $4.1\%$ increase in the part segmentation task.
\end{itemize}

\section{Related Work}

\subsection{Sampling Methods for Point Cloud}

\begin{figure*}[t]
    \centering
    \includegraphics[width=0.9\linewidth]{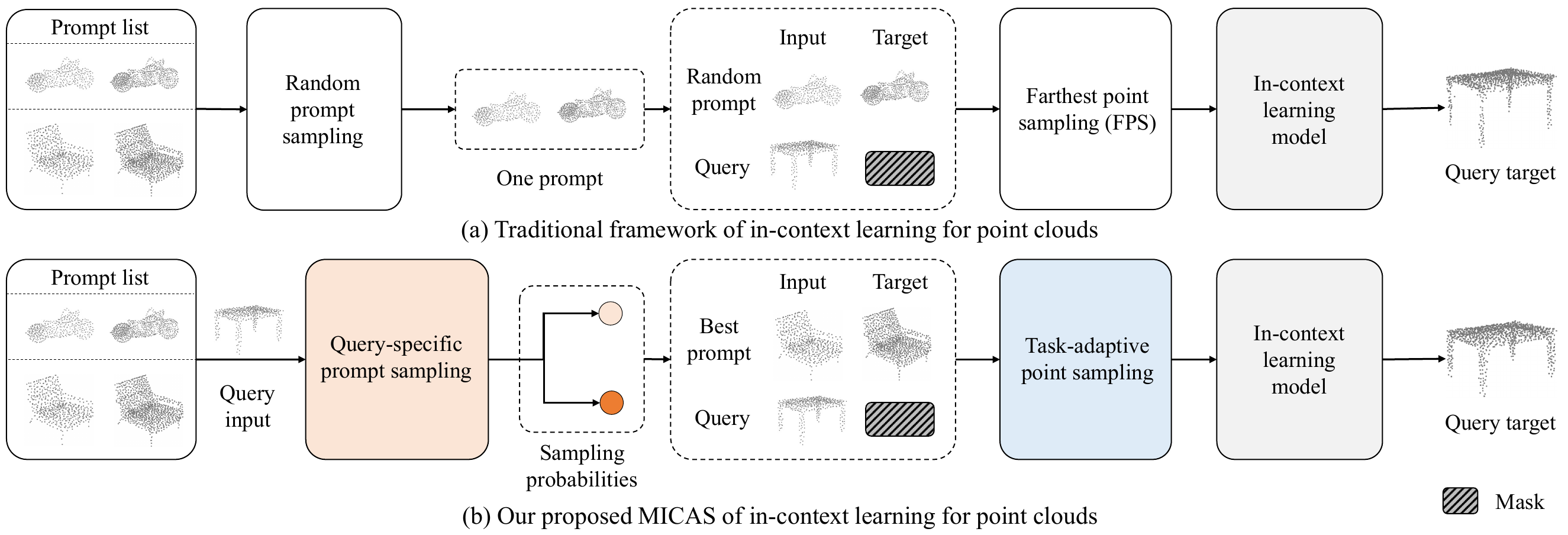}
    \vspace{-0.6em}
    \caption{Comparison between the proposed MICAS and the traditional in-context learning framework.}
    \label{overview}
    \vspace{-1.2em}
\end{figure*}

Point cloud sampling is essential for representing object shape and topology efficiently, enabling large-scale point cloud analysis~\cite{wen2023learnable, li2018pointcnn, liu2019relation, ma2022rethinking}. Existing methods can be categorized into mathematical statistics-based and learnable task-based approaches. First, mathematical statistics-based methods\cite{hu2020randla, thomas2019kpconv, chen2021unsupervised, qi2017pointnet++, li2018pointcnn, groh2018flex} are task-agnostic, leveraging structural and geometric properties. Techniques include random sampling\cite{hu2020randla}, grid sampling~\cite{thomas2019kpconv}, farthest point sampling (FPS)~\cite{qi2017pointnet++, li2018pointcnn}, and Inverse Density Importance Sampling (IDIS)~\cite{groh2018flex}. While effective, these methods overlook task-specific information. Second, learnable task-based methods~\cite{dovrat2019learning, lang2020samplenet, yan2020pointasnl, wen2023learnable} design sampling networks tailored to specific tasks and guided by task losses. Dovrat~\etal\cite{dovrat2019learning} propose S-Net, a learnable network that generates point subsets and enhances them with ProgressiveNet, which prioritizes task-relevant points. SampleNet~\cite{lang2020samplenet} introduces differentiable sampling using weighted averages of nearest neighbors, while IndexSample~\cite{wu2021indexsample} improves results with a confidence layer. SkeletonNet~\cite{wen2023learnable} uses Gumbel-softmax for discrete sampling, and CP-Net~\cite{nezhadarya2020adaptive} performs adaptive down-sampling. PAT~\cite{yang2019modeling} employs group shuffle attention, PointASNL~\cite{yan2020pointasnl} adaptively adjusts point features and local normalization, Pra-net~\cite{cheng2021net} integrates intra-region structure learning for local adaptation, and APES~\cite{wu2023attention} utilizes edge detection for adaptive sampling.

However, existing mathematical statistics-based sampling overlooks information from both the point cloud and the task. 
Meanwhile, existing learnable task-based sampling focuses on inter-point cloud adaptation within the same task, neglecting inter-task adaptation within the same point cloud. 
To handle this issue, we propose task-adaptive point sampling to leverage task-specific information from prompts for customized, efficient sampling across tasks. 

\subsection{Demonstration Retrieval for ICL}
The sensitivity of in-context learning to demonstration selection~\cite{xu2024context} has led to the development of various retrieval techniques, categorized into similarity-based and diversity-based methods. First, similarity-based retrieval assumes that demonstrations resembling the query provide valuable guidance~\cite{liu2021makes}. Methods like KATE~\cite{liu2021makes} retrieve semantically similar examples to construct prompts, while EPR~\cite{rubin2021learning} uses similarity scores based on inner products. PARC~\cite{nie2022cross} enriches contexts with semantically similar sentences, and UDR~\cite{li2023unified} introduces a multi-task list-wise ranking framework to mine high-quality demonstrations. Second, diversity-based retrieval focuses on reducing redundancy, providing varied perspectives, and ensuring query coverage~\cite{zhang2022automatic, yu2022generate, levy2022diverse}. Auto-CoT~\cite{zhang2022automatic} diversifies sampled questions to construct reasoning chains, GENREAD~\cite{yu2022generate} uses clustering to synthesize prompts from diverse clusters, and Cover-LS~\cite{levy2022diverse} selects demonstrations that ensure structural coverage for better generalization.

Drawing inspiration from existing demonstration retrieval methods, we implement a simple yet effective probability-based retrieval approach of 3D point cloud, introducing a novel query-specific prompt sampling module.

\subsection{Point Cloud in Large Language Model Era}
With the rapid advancements in Large Language Models (LLMs)\cite{radford2021learning, touvron2023llama}, various point cloud methods integrating LLMs have emerged. For example, PointCLIP\cite{zhang2022pointclip} and CLIP2~\cite{zeng2023clip2} align 3D data with language representations, leveraging multi-view depth maps and few-shot fine-tuning, triplet proxies collection scheme and cross-modal pretraining, respectively. MiniGPT-3D~\cite{tang2024minigpt} efficiently aligns 3D data with LLMs using 2D priors. Point-E~\cite{nichol2022point} generates 3D point clouds from prompts, and PointBind~\cite{guo2023point} offers a unified framework for 3D multi-modal tasks. PointLLM~\cite{xu2025pointllm} and SegPoint~\cite{he2025segpoint} utilize LLaMA~\cite{touvron2023llama} for understanding point clouds, while PIC~\cite{fang2024explore} and DG-PIC~\cite{jiang2025dg} apply ICL for multi-task and multi-domain point cloud processing.

In this work, we find the inter-task and intra-task sensitivity issues in current ICL methods of point clouds, stemming from inflexible sampling strategies that lack context adaptation at both the point and prompt levels. To address these challenges, we propose an enhanced ICL method with a multi-grained adaptive sampling mechanism.

\section{Methodology}

\subsection{Preliminaries}

\textbf{Problem Settings.} 
We formally define the problem settings for in-context learning with 3D point clouds.
As illustrated in Figure~\ref{overview} (a), each input sample comprises two pairs of ``input-target'' point clouds, similar to the setup used in 2D-context learning~\cite{fang2024explore}. 
One pair serves as a prompt, and the other pair serves as a query. Each pair consists of an input point cloud and its corresponding output point cloud for the given task~\cite{xu2024context, fang2024explore}. The prompts represent four typical PCP tasks: reconstruction~\cite{jenke2006bayesian, mandikal2019dense}, denoising~\cite{luo2021score, javaheri2017subjective}, registration~\cite{yang2020teaser, qi2017pointnet}, and part segmentation~\cite{shao2022active, landrieu2018large}. Following established protocols~\cite{fang2024explore, liu2024point}, the network is trained to reconstruct randomly masked parts of the ``target'' point cloud in both the prompt and the query. 
During inference, the model reconstructs the ``target'' point cloud of the query.

\noindent
\textbf{Revisiting Point Cloud In-Context Learning Model.} 
Before presenting our method, we formally introduce the framework of in-context learning for point clouds.
Using the pioneering work PIC~\cite{fang2024explore} as an example, which introduces a new benchmark, the framework comprises data processing, model design, and model training.

Regarding data processing, PIC~\cite{fang2024explore} begins by considering two pairs of ``input-target'' point clouds, denoted as query $Q=(X_q, Y_q)$ and prompt $P=(X_p, Y_p)$. 
It first applies  Farthest Point Sampling (FPS)~\cite{qi2017pointnet++} to select $N$ central points $C_{X_q}$ and $C_{X_p}$ from $X_q$ and $X_p$, respectively. 
To ensure alignment between the sampled central points derived from the ``input'' and ``target'' point clouds, a Joint Sampling (JS) module is employed.
This module uses the point indexes of central points $C_{X_q}$ and $C_{X_p}$ to locate the corresponding position points in the ``target'' point clouds ${Y_q}$ and ${Y_p}$ as their center points $C_{Y_q}$ and $C_{Y_p}$, respectively. 
Subsequently, the K-Nearest Neighbors (KNN)\cite{fix1985discriminatory} technique transforms $(X_q, Y_q, X_p, Y_p)$ into $N$ point patches $(R_{X_q}, R_{Y_q}, R_{X_p}, R_{Y_p})$ based on these central points, which are then encoded into tokens.
Finally, these point patches are encoded into tokens.

In model design, PIC~\cite{fang2024explore} adopts a mask-point modeling (MPM) strategy with a transformer-based encoder-decoder architecture.
A $1 \times 1$ convolutional layer serves as the task head for reconstructing the point clouds.

During model training, PIC~\cite{fang2024explore} utilizes two pairs of point patches, the query point patches $(R_{X_q}, R_{Y_q})$ and the prompt point patches $(R_{X_p}, R_{Y_p})$, to perform a masked point reconstruction task. 
It first randomly masks the point patches within $R_{Y_q}$ and $R_{Y_p}$ and then trains the model using the Chamfer Distance~\cite{fan2017point} loss $\mathcal{L}_{cd}$, defined as:
\begin{equation}
  \label{eq:cd}
  \begin{aligned}
    \mathcal{L}_{cd}(R_{pred}, G) &= \frac{1}{|R_{pred}|}\sum_{r\in R_{pred}}\min_{g\in G}||r-g||^2_2 \\
    & + \frac{1}{|G|}\sum_{g\in G}\min_{r\in R_{pred}}||g-r||^2_2,
  \end{aligned}
\end{equation}
where $\mathcal{L}_{cd}$ measures the discrepancy between each predicted patch $R_{pred}$ and its corresponding ground truth patch $G$, $|R_{pred}|$ and $|G|$ represent the number of points in patch $R_{pred}$ and patch $G$, respectively. 
During inference, the model predicts the entire masked ``target'' point cloud $Y_q$ for the query, which is shown in Figure~\ref{overview} (a).

However, PIC~\cite{fang2024explore} employs Farthest Point Sampling (FPS), which lacks context adaptation at both the point and prompt levels, leading to sensitivity issues across and within tasks. As shown in Figure~\ref{intro} and Table~\ref{tab:ablation}, FPS often selects noisy points as center points in the denoising task, causing the model’s CD loss to remain high. 
To overcome these critical limitations, we propose a novel Multi-grained In-Context Adaptive Sampling mechanism, dubbed MICAS, which fundamentally rethinks point cloud in-context learning by incorporating task-adaptive point sampling and query-specific prompt sampling. 
This new approach significantly enhances the adaptability and robustness of point cloud processing tasks shown in Figure~\ref{presentation_all} and Table~\ref{tab:ablation}, addressing the inter-task and intra-task sensitivity issues that previous methods, such as PIC, fail to resolve.

\begin{figure*}[t]
  \centering
  \includegraphics[width=1.0\linewidth]{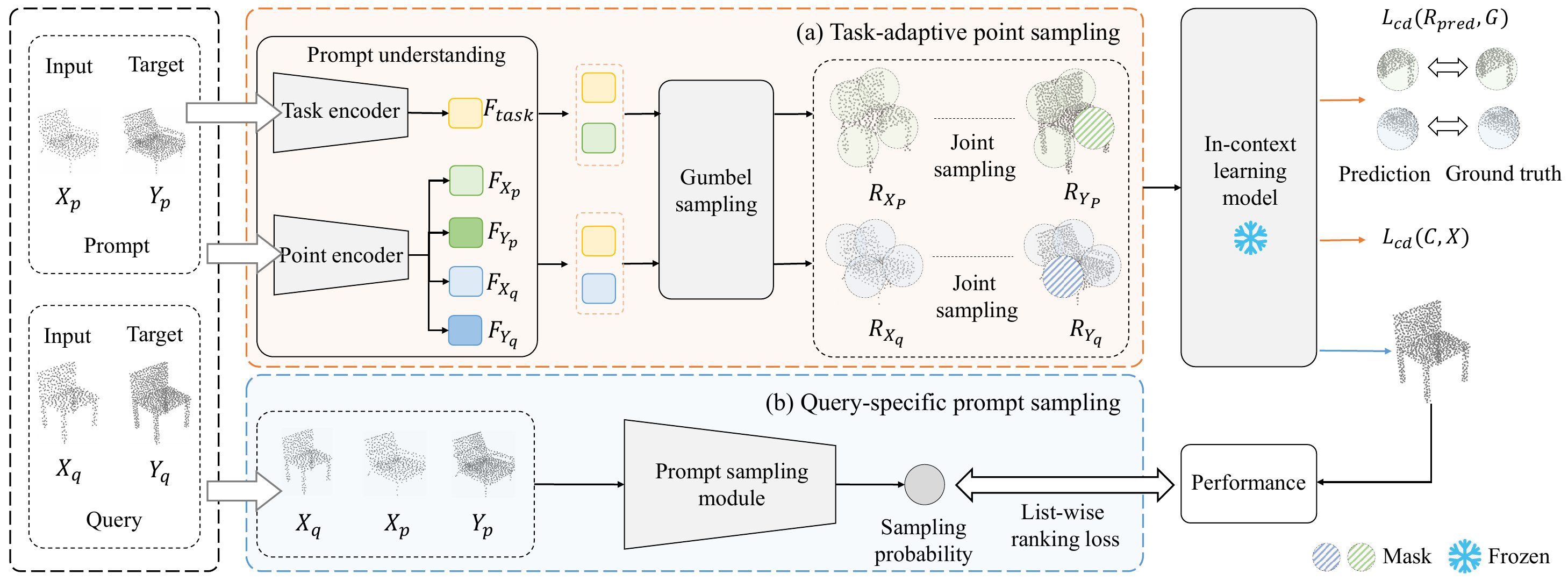}
  \vspace{-1.0em}
  \caption{Overview of the proposed MAL-ICL. (a) Task-adaptive point sampling is designed to achieve better point-level sampling. (b) Query-specific prompt sampling aims to infer the most effective prompt-level sampling.}
  \label{framework}
  \vspace{-1.0em}
\end{figure*}

\subsection{Task-adaptive Point Sampling}
As illustrated in Figure~\ref{framework} (b), we introduce a task-adaptive point sampling module to address the \textit{inter-task sensitivity} (cf. Figure~\ref{intro} (a)) by focusing on understanding and applying task information from prompts during the sampling stage. This module comprises two key components: prompt understanding, which extracts relevant task features and point features, and Gumbel sampling, which achieves differentiable sampling via the Gumbel-softmax~\cite{jang2016categorical, yang2019modeling} leveraging these extracted dual-level features. 

\textbf{1) Prompt Understanding.} To accurately understand the point cloud information in the ``input-target'' point clouds of query $Q=(X_q, Y_q)$ and prompt $P=(X_p, Y_p)$, we adopt PointNet~\cite{qi2017pointnet} as our task encoder and point encoder, as shown in Figure~\ref{framework} (a).
First, we employ a task encoder that incorporates the max pooling layer and previous from the PointNet classification branch~\cite{qi2017pointnet}, enabling it to extract task-relevant information from prompts.
Its objective is to process the prompt $P=(X_p, Y_p)$ and generate the corresponding task feature $F_{task}$. 
Specifically, we concatenate the prompt $P=(X_p, Y_p)$, and feed this concatenation into the task encoder $\Phi_{task}$ to yield the task feature $F_{task}$:
\begin{equation}
  \begin{aligned}
    F_{task} = \Phi_{task}(X_p \oplus Y_p),
  \end{aligned}
\end{equation}
where $\oplus$ denotes concatenation operation.

Second, to extract point feature information from each point cloud, we employ a point encoder based on the PointNet segmentation branch~\cite{qi2017pointnet}. 
Its purpose is to process any given point cloud $X_*$ and produce the associated point features $F_{X_*}$:
\begin{equation}
  \begin{aligned}
    F_{X_*} = \Phi_{point}(X_*),
  \end{aligned}
\end{equation}
where $X_*$ refers to any of the point clouds $X_q$, $Y_q$, $X_p$, and $Y_p$. Accordingly, $F_{X_*}$ represents the features of point clouds, namely $F_{X_q}$, $F_{Y_q}$, $F_{X_p}$, and $F_{Y_p}$, respectively.

\textbf{2) Gumbel Sampling.} 
We utilize the task feature $F_{task}$ and the point features  $F_{X_q}$ and $F_{X_p}$  to achieve differentiable sampling by integrating the Gumbel-softmax approach~\cite{jang2016categorical, yang2019modeling}. 
This method performs a ``soft'' selection that mimics one-hot encoding by blending probabilities rather than making a hard selection of a single point.

As illustrated in Figure~\ref{framework} (a), we initially merge the task feature $F_{task} \in \mathbb{R}^{d1}$ with the point feature $F_{X_q} \in \mathbb{R}^{S\times d2}$ to create enhanced point features $\hat{F}_{X_q} \in \mathbb{R}^{S\times(d1+d2)}$:
\begin{equation}
  \begin{aligned}
    \hat{F}_{X_q} = F_{task} \oplus F_{X_q},
  \end{aligned}
\end{equation}
where $d1$ and $d2$ represent the feature dimensions, and $S$ denotes the number of points in the point cloud. 
Then, the enhanced point features $\hat{F}_{X_q}$ are passed through a fully connected layer with weight parameter $W\in \mathbb{R}^{(d1+d2)\times N}$ to yield sampling weights $SW\in \mathbb{R}^{S\times N}$:
\begin{equation}
  \begin{aligned}
    SW = \hat{F}_{X_q} \times W,
  \end{aligned}
\end{equation}
where $N$ indicates the number of selected points.

Subsequently, the sampling weight $SW$ is normalized using the Gumbel-softmax~\cite{jang2016categorical, yang2019modeling}, which employs a discrete reparameterization technique to obtain smooth gradients by continuously relaxing the categorical variable~\cite{wen2023learnable}. 
Given the Gumbel noise $g = (g_1, \dots, g_k)$, where each $g_i$ is independently drawn from a Gumbel distribution within 0 and 1, the soft sampling weight $SW_{gs}$ is calculated as:
\begin{equation}
  \begin{aligned}
    SW_{gs} = \textit{softmax} \left( (\log(SW) + g) / {\tau} \right),
  \end{aligned}
\end{equation}
where $\tau > 0$ is the annealing temperature, and the softmax function operates along the dimension of points. 
The Gumbel-softmax mechanism ensures that the newly generated points remain within the three-dimensional space of the original point cloud. 
Ultimately, the selected central points $C_{X_q}\in \mathbb{R}^{N\times 3}$ are generated by projecting the sampling weight $SW_{gs}$ onto the original point cloud $X_q\in R^{S\times 3}$:
\begin{equation}
  \begin{aligned}
    C_{X_q} = SW_{gs}^T \times X_q.
  \end{aligned}
\end{equation}
The same process is applied to derive the sampling points  $C_{X_p}$ for the point cloud  $X_p$.
Following the methodology of Fang \etal~\cite{fang2024explore}, we employ Joint Sampling and KNN techniques to produce $N$ point patches $(R_{X_q}, R_{Y_q}, R_{X_p}, R_{Y_p})$, which are then input into the in-context learning model (\eg, PIC~\cite{fang2024explore}) for masked point modeling.

\textbf{3) Loss Function.} To enhance the training of the task-adaptive point sampling module, we implement an additional $\mathcal{L}_{cd}(C, X)$ loss function based on Equation~\ref{eq:cd}. This new loss function quantifies the discrepancy between the sampled central points $C$ and the original point cloud $X$, as shown in Figure~\ref{framework}. Finally, the training loss of the task-adaptive point sampling module, denoted as $\mathcal{L}_{sampling}$, is defined as follows:
\begin{equation}
  \label{eq:sampling}
  \begin{aligned}
    \mathcal{L}_{sampling} = \mathcal{L}_{cd}(R_{pred}, G) + \alpha \cdot \mathcal{L}_{cd}(C, X),
  \end{aligned}
\end{equation}
where $R_{pred}$ and $G$ respectively represent the predicted patch and its ground-truth patch, as introduced in Equation~\ref{eq:cd}.
The hyperparameter $\alpha$ is to modulate the influence of the CD loss between the sampled points and the original point cloud.

\subsection{Query-specific Prompt Sampling}
To address the \textit{intra-task sensitivity} issue depicted in Figure~\ref{intro} (c), we { introduce a query-specific prompt sampling module designed to select the most suitable prompt, as depicted in Figure~\ref{framework} (b).}

\textbf{1) Pseudo Label.} 
Inspired by UDR~\cite{li2023unified}, which addresses prompt retrieval in natural language processing, we collect training examples for our prompt sampling module by utilizing the output signals from the in-context learning model $\Phi_{ICL}$ (\eg, PIC~\cite{fang2024explore}).
Specifically, given a query ``input'' point cloud $X_q$ and a prompt $P=(X_p, Y_p)$, $\Phi_{ICL}$ processes these inputs to generate a predicted query ``target'' point cloud $\tilde{Y}_q$. 
We then evaluate the performance by comparing $\tilde{Y}_q$ with the ground-truth query ``target'' point cloud $Y_q$, using metrics such as CD loss or mIOU. 
The resulting performance serves as the pseudo label $\tilde{y}$ for the training of the prompt sampling module, as illustrated in Figure~\ref{framework}:
\begin{equation}
  \label{eq:pseudo_label}
  \begin{aligned}
    \tilde{y} = \Phi_{ICL}(X_q, X_p, Y_p).
  \end{aligned}
\end{equation} 
To ensure consistency across different tasks, we employ max-min normalization~\cite{kim2021normalization, vafaei2022assessing}. 
This normalization maintains the maximum and minimum performance values for each task, allowing us to normalize performance indicators across different tasks to the range $[0, 1]$.

\textbf{2) Sampling Probability.} Our goal is to utilize the query ``input'' point cloud to generate a sampling probability for each candidate prompt. 
Specifically, we first combine the query ``input'' point cloud $X_q$ with the prompt $P=(X_p, Y_p)$ to form a fused point cloud $\tilde{X} \in \mathbb{R}^{3\times S\times 3}$:
\begin{equation}
  \label{eq:fusion}
  \begin{aligned}
    \tilde{X} = (X_q \oplus X_p \oplus Y_p),
  \end{aligned}
\end{equation}
where $\oplus$ denotes the concatenation along the point dimension, and $S$ represents the number of points in the point cloud. 
We randomly select $K$ prompts for each query ``input'' point cloud,  generating $K$ new point clouds $\tilde{X}_i$. 
These $\tilde{X}_i$ are then passed through the prompt sampling module $\Phi_{prompt}$ \footnote{The prompt sampling module is model-agnostic. In this paper, we employ PointNet~\cite{qi2017pointnet} as the prompt sampling module.} to produce $K$ sampling probabilities $Prob = \{prob_1, prob_2, \dots, prob_K\}$, where each $prob_i$ is defined as:
\begin{equation}
  \label{eq:prob}
  \begin{aligned}
    prob_i = \Phi_{prompt}(\tilde{X}_i).
  \end{aligned}
\end{equation}

\textbf{3) Loss Function.} Given a query ``input'' point cloud and $K$ randomly selected prompts, we first generate $K$ pseudo labels $\tilde{Y} = \{\tilde{y}_1, \tilde{y}_2, \dots, \tilde{y}_K\}$ using Equation~\ref{eq:pseudo_label}. Then, we compute $K$ sampling probabilities $Prob = \{prob_1, prob_2, \dots, prob_K\}$ by employing Equation~\ref{eq:prob}. Finally, we utilize the list-wise ranking loss $\mathcal{L}_{listwise\_rank}$ to evaluate and optimize ranking orders~\cite{burges2010ranknet, li2023unified, xu2024context}, as shown in Figure~\ref{framework} (b).
\begin{equation}
  \label{eq:retrieval}
  \begin{aligned}
\mathcal{L}_{listwise\_rank} & =\sum_{i,j}^{K}\textit{max}(0,\frac{1}{r(\tilde{y}_i)}-\frac{1}{r(\tilde{y}_j)}) \\
& \times \textit{log}(1+e^{(prob_j-prob_i)}),
\end{aligned}
\end{equation}
where $r(\tilde{y})$ indicates the ranking order of $\tilde{y}$ among these candidate prompts.

During inference, given a query ``input'' $X_q$, we first use the prompt sampling module to select the best prompt with the highest probability among $K$ candidates, as shown in Figure~\ref{overview} (b).
Then, we input $X_q$ and selected prompt into PIC~\cite{fang2024explore} to predict the ``target'' point cloud for the query.

\subsection{Model Training}
\label{sec:model_training}

Task-adaptive point sampling learns each prompt individually, whereas query-specific prompt sampling evaluates multiple prompts simultaneously.
Jointly training these two modules could increase the learning complexity of task-adaptive point sampling, slow convergence, and create unnecessary entanglement between the modules. 
Adopting a step-wise training strategy, as suggested in previous studies~\cite{moryossef2019step, bolukbasi2017adaptive}, can simplify the problem, improve robustness, and make the learning process more manageable. 
Therefore, we employ this strategy for our proposed MICAS.

First, we train the task-adaptive point sampling module, replacing the central points typically selected by FPS with those produced by our sampling method. This phase focuses on optimizing point sampling and uses the Chamfer Distance (CD) loss (cf. Equation~\ref{eq:sampling}), while the query-specific prompt sampling module remains inactive.

Once the task-adaptive point sampling module is trained and its parameters are fixed, we proceed to train the query-specific prompt sampling module. 
This module  analyzes each query and its candidate prompts to predict sampling probabilities, rank them, and optimize using the list-wise ranking loss (cf. Equation~\ref{eq:retrieval}).

\section{Experiments}

\begin{table*}[]
  \centering
  \caption{Comparison with state-of-the-art models on the ShapeNet In-Context~\cite{fang2024explore}. For reconstruction, denoising, and registration, we report Chamfer Distance (CD)~\cite{fan2017point} loss (x1000). For part segmentation, we report mIOU. Copy: uses the prompt's ``target'' point cloud as its prediction. The \blue{blue} and \underline{underline} values indicate the best and second-best results.}
  \scriptsize
  \setlength\tabcolsep{1.2mm}
  \renewcommand\arraystretch{1.3}
  \begin{tabular}{l|c|cccccc|cccccc|cccccc|c}
  \toprule[0.15em]
  \multirow{2}{*}{Models} & \multirow{2}{*}{Venues} & \multicolumn{6}{c|}{Reconstruction CD $\downarrow$}        & \multicolumn{6}{c|}{Denoising CD $\downarrow$}           & \multicolumn{6}{c|}{Registration CD $\downarrow$}     & Part Seg.  \\
  \multicolumn{1}{c|}{}    & \multicolumn{1}{c|}{}                  
  & L1    & L2    & L3    & L4    & L5    & \multicolumn{1}{c|}{Avg.} 
  & L1    & L2    & L3    & L4    & L5    & \multicolumn{1}{c|}{Avg.}  
  & L1    & L2    & L3    & L4    & L5    & \multicolumn{1}{c|}{Avg.}  
  & mIOU$\uparrow$              \\ \hline
  \multicolumn{21}{c}{Task-specific models (trained separately)}           \\ \hline
  \multicolumn{1}{l|}{PointNet~\cite{qi2017pointnet}} & \multicolumn{1}{c|}{CVPR'17}   
  & 3.7   & 3.7   & 3.8   & 3.9   & 4.1   & \multicolumn{1}{c|}{3.9} 
  & 4.1   & 4.0   & 4.1   & 4.0   & 4.2   & \multicolumn{1}{c|}{4.1}  
  & 5.3   & 5.9   & 6.9   & 7.7   & 8.5   & \multicolumn{1}{c|}{6.9}  
  & 77.5      \\

  \multicolumn{1}{l|}{DGCNN~\cite{wang2019dynamic}}  & \multicolumn{1}{c|}{TOG'19}  
  & 3.9   & 3.9   & 4.0   & 4.1   & 4.3   & \multicolumn{1}{c|}{4.0}   
  & 4.7   & 4.5   & 4.6   & 4.5   & 4.7   & \multicolumn{1}{c|}{4.6}   
  & 6.2   & 6.7   & 7.3   & 7.4   & 7.7   & \multicolumn{1}{c|}{7.1}   
  & 76.1   \\
 
  \multicolumn{1}{l|}{PCT~\cite{guo2021pct}}  & \multicolumn{1}{c|}{CVM'21}  
  & 2.4   & 2.4   & 2.5   & 2.6   & 3.0   & \multicolumn{1}{c|}{2.6}  
  & 2.3   & 2.2   & 2.2   & 2.2   & 2.3   & \multicolumn{1}{c|}{2.2}  
  & 5.3   & 5.7   & 6.3   & 6.9   & 7.2   & \multicolumn{1}{c|}{6.3}   
  & 79.5      \\

  \multicolumn{1}{l|}{ACT~\cite{dong2022autoencoders}}  & \multicolumn{1}{c|}{ICLR'23}  
  & 2.4   & 2.5   & 2.3   & 2.5   & 2.8   & \multicolumn{1}{c|}{2.5}   
  & 2.2   & 2.3   & 2.2   & 2.3   & 2.5   & \multicolumn{1}{c|}{2.3}   
  & 5.1   & 5.6   & 5.9   & 6.0   & 7.0   & \multicolumn{1}{c|}{5.9}   
  & 81.2      \\ \hline
  \multicolumn{21}{c}{Multi-task models: share backbone + multi-task heads}               \\ \hline
  \multicolumn{1}{l|}{PointNet~\cite{qi2017pointnet}}   & \multicolumn{1}{c|}{CVPR'17}  
  & 87.2  & 86.6  & 87.3  & 90.8  & 92.2  & \multicolumn{1}{c|}{88.8} 
  & 17.8  & 22.0  & 25.6  & 30.4  & 33.2  & \multicolumn{1}{c|}{25.8}  
  & 25.4  & 22.6  & 24.9  & 25.7  & 26.9  & \multicolumn{1}{c|}{25.1} 
  & 15.3      \\
  \multicolumn{1}{l|}{DGCNN~\cite{wang2019dynamic}}  & \multicolumn{1}{c|}{TOG'19}  
  & 38.8  & 36.6  & 37.5  & 37.9  & 42.9  & \multicolumn{1}{c|}{37.7}  
  & 6.5   & 6.3   & 6.5   & 6.4   & 7.1   & \multicolumn{1}{c|}{6.5}  
  & 12.5  & 14.9  & 17.9  & 19.7  & 20.7  & \multicolumn{1}{c|}{17.1}  
  & 17.0      \\
  \multicolumn{1}{l|}{PCT~\cite{guo2021pct}}  & \multicolumn{1}{c|}{CVM'21}   
  & 34.7  & 44.1  & 49.9  & 50.0  & 52.3  & \multicolumn{1}{c|}{46.2}  
  & 11.2  & 10.3  & 10.7  & 10.2  & 10.5  & \multicolumn{1}{c|}{10.6} 
  & 24.4  & 26.0  & 29.6  & 32.8  & 34.7  & \multicolumn{1}{c|}{29.5} 
  & 16.7      \\ 
  \multicolumn{1}{l|}{Point-MAE~\cite{pang2022masked}} & \multicolumn{1}{c|}{ECCV'22}  
  & 5.5  & 5.5  & 6.1  & 6.4  & 6.4  & \multicolumn{1}{c|}{6.0}  
  & 5.6  & 5.4  & 5.6  & 5.5  & 5.8  & \multicolumn{1}{c|}{5.6} 
  & 11.4  & 12.8  & 14.8  & 16.0  & 16.9  & \multicolumn{1}{c|}{14.5} 
  & 5.4      \\ 
  \multicolumn{1}{l|}{ACT~\cite{dong2022autoencoders}} & \multicolumn{1}{c|}{ICLR'23}  
  & 7.4  & 6.6  & 6.5  & 6.6  & 7.0  & \multicolumn{1}{c|}{6.8}  
  & 7.3  & 6.8  & 7.0  & 6.8  & 7.2  & \multicolumn{1}{c|}{7.0}  
  & 12.2  & 14.4  & 19.4  & 25.5  & 29.0  & \multicolumn{1}{c|}{20.1} 
  & 12.1      \\ 
  \multicolumn{1}{l|}{I2P-MAE~\cite{zhang2023learning}} & \multicolumn{1}{c|}{CVPR'23}  
  & 17.0  & 16.0  & 16.7  & 17.2  & 18.5  & \multicolumn{1}{c|}{17.2} 
  & 20.6  & 20.4  & 20.1  & 18.3  & 18.8  & \multicolumn{1}{c|}{19.6} 
  & 32.5  & 31.3  & 31.1  & 31.6  & 31.2  & \multicolumn{1}{c|}{31.5} 
  & 22.6      \\ 
  \multicolumn{1}{l|}{ReCon~\cite{qi2023contrast}} & \multicolumn{1}{c|}{ICML'23} 
  & 12.4  & 12.1  & 12.4  & 12.5  & 13.1  & \multicolumn{1}{c|}{12.5}  
  & 20.4  & 24.5  & 27.2  & 29.2  & 32.5  & \multicolumn{1}{c|}{26.9}  
  & 14.7  & 16.3  & 19.2  & 21.5  & 22.5  & \multicolumn{1}{c|}{18.8} 
  & 7.7      \\ \hline
  \multicolumn{21}{c}{In-context learning models}                     \\ \hline
  \multicolumn{1}{l|}{Copy}      & \multicolumn{1}{l|}{}              
  & 155 & 153 & 152 & 156 & 155 & \multicolumn{1}{c|}{154} 
  & 149 & 155 & 157 & 155 & 155 & \multicolumn{1}{c|}{154} 
  & 155 & 157 & 156 & 148 & 154 & \multicolumn{1}{c|}{154}
  & 24.2    \\
  \multicolumn{1}{l|}{Point-BERT~\cite{yu2022point}} & \multicolumn{1}{c|}{CVPR'22}  
  & 288 & 285 & 292 & 286 & 308 & \multicolumn{1}{c|}{292} 
  & 292 & 293 & 298 & 296 & 299 & \multicolumn{1}{c|}{296} 
  & 291 & 295 & 294 & 295 & 298 & \multicolumn{1}{c|}{294}
  & 0.7        \\
  
    \rowcolor{mygray}
    \multicolumn{1}{l|}{PIC-Cat~\cite{fang2024explore}}    & \multicolumn{1}{c|}{NeurIPS'23}    
    & \textbf{\textcolor{blue}{3.2}}   & \textbf{\textcolor{blue}{3.6}}   & 4.6   & 4.9   & \textbf{\textcolor{blue}{5.5}}   &\multicolumn{1}{c|}{\textbf{\textcolor{blue}{4.3}}}  
    & \textbf{\textcolor{blue}{3.9}}  & \underline{4.6}   & 5.3   & 6.0   & 6.8   & \multicolumn{1}{c|}{5.3}  
    & 10.0   & 11.4   & 13.8  & 16.9  & 18.6 & \multicolumn{1}{c|}{14.1} 
    & 79.0     \\ 
      
    \multicolumn{1}{l|}{PIC-Sep~\cite{fang2024explore}}    & \multicolumn{1}{c|}{NeurIPS'23}    
    & 4.7   & 4.3   & \underline{4.3}   & \textbf{\textcolor{blue}{4.4}}   & 5.7   &\multicolumn{1}{c|}{4.7}  
    & 6.3  & 7.2   & 7.9   & 8.2   & 8.6   & \multicolumn{1}{c|}{7.6}   
    & 8.6   & 9.2   & 10.2  & 11.3  & 12.4 & \multicolumn{1}{c|}{10.3} 
    & 75.0     \\ 
  
  \rowcolor{mygray}
  \multicolumn{1}{l|}{PIC-S-Cat~\cite{liu2024point}}    & \multicolumn{1}{c|}{Arxiv'24}   
  & 9.3   &  5.1   &  4.8   &  5.0   &  10.3   & \multicolumn{1}{c|}{6.9}  
  & 4.7  & 5.7   & 6.5   &  7.4   & 8.2   &  \multicolumn{1}{c|}{6.5}   
  & 12.8   & 15.8   &  23.9  & 31.2  & 36.9 & \multicolumn{1}{c|}{24.1}  
  & 83.8     \\ 
  
  \multicolumn{1}{l|}{PIC-S-Sep~\cite{liu2024point}}    & \multicolumn{1}{c|}{Arxiv'24}    
  & 4.6   & 4.5   &  4.5   & 4.8   & 7.1   &\multicolumn{1}{c|}{5.1}   
  & 9.4  & 11.7   &  12.5   & 13.1   &  13.4   & \multicolumn{1}{c|}{12.0}   
  & 6.0   &  \underline{6.1}   & \underline{7.6}  & \underline{6.7}  & \underline{7.3} &  \multicolumn{1}{c|}{\underline{6.7}}  
  & 83.7     \\ 
  
  \rowcolor{mygray}
  \multicolumn{1}{l|}{PIC-Cat~\cite{fang2024explore} + MICAS}     & \multicolumn{1}{c|}{Ours}  
  & 4.6 & 4.2 & 4.5 & 4.8 & 5.7 & \multicolumn{1}{c|}{4.7} 
  & \underline{4.2} & \textbf{\textcolor{blue}{4.4}} & \textbf{\textcolor{blue}{4.6}} & \textbf{\textcolor{blue}{4.9}} & \textbf{\textcolor{blue}{5.1}} & \multicolumn{1}{c|}{\textbf{\textcolor{blue}{4.6}}} 
  & \underline{5.7} & 6.5 & 9.1 & 12.5 & 15.4 & \multicolumn{1}{c|}{9.8} 
  & \textbf{\textcolor{blue}{87.9}} \\ 
  
  \multicolumn{1}{l|}{PIC-Sep~\cite{fang2024explore} + MICAS}     & \multicolumn{1}{c|}{Ours} 
  & \underline{3.8} & \underline{3.9} & \textbf{\textcolor{blue}{4.0}} & \textbf{\textcolor{blue}{4.4}} & \underline{5.6} & \multicolumn{1}{c|}{\textbf{\textcolor{blue}{4.3}}} 
  & 4.4 & 4.9 & \underline{5.2} & \underline{5.5} & \underline{5.7} & \multicolumn{1}{c|}{\underline{5.1}} 
  & \textbf{\textcolor{blue}{3.4}} & \textbf{\textcolor{blue}{3.6}} & \textbf{\textcolor{blue}{3.7}} & \textbf{\textcolor{blue}{3.8}} & \textbf{\textcolor{blue}{4.0}} & \multicolumn{1}{c|}{\textbf{\textcolor{blue}{3.7}}} 
  &  \underline{86.8} \\ 
  
  \bottomrule[0.1em]
  \end{tabular}
  \vspace{-1.0em}
  \label{tab:sota}
  \end{table*}

\subsection{Experimental Settings}
\textbf{Dataset.} The proposed MICAS is rigorously evaluated using the ShapeNet In-Context Dataset, introduced in the PIC~\cite{fang2024explore}. 
This dataset comprises ``input-target'' point cloud pairs, each derived from well-known repositories such as ShapeNet~\cite{chang2015shapenet} and ShapeNetPart~\cite{yi2016scalable}. 
The ``input'' point cloud serves the task query, while the ``target'' represents the expected outcome. The dataset is extensive, featuring $174,404$ samples for training and $43,050$ for testing, across four distinct tasks: registration, reconstruction, denoising, and part segmentation. 
Each task is divided into five levels of difficulty to assess model performance comprehensively.

\noindent
\textbf{Evaluation Metrics.} We employ the Chamfer Distance (CD)~\cite{fan2017point} and Mean Intersection over Union (mIOU) as the primary evaluation metrics for different tasks. For registration, reconstruction, and denoising tasks, CD is used to measure the structural discrepancy between the predicted and ground-truth point clouds. For part segmentation, mIOU is utilized to appraise segmentation performance.

\noindent
\textbf{Implementation Details.} Following PIC~\cite{fang2024explore}, we sample $1,024$ points from each point cloud and segment them into $64$ patches, each containing $32$ neighboring points. PointNet~\cite{qi2017pointnet} is used as the task encoder, point encoder, and prompt sampling module (cf. Figure~\ref{framework}). For task-adaptive point sampling, we set the initial learning rate to $0.0001$, reducing it to $1e-6$ over $60$ epochs using a Cosine Annealing Scheduler~\cite{loshchilov2016sgdr}, with a batch size of $72$ and a sampling loss hyperparameter $\alpha$ of $0.5$. For query-specific prompt sampling, $8$ candidate prompts are randomly selected per query, with a learning rate of $0.00001$, decay to $0.000001$, $30$ training epochs, and a batch size of $9$.

\noindent
\textbf{Model Variants.} PIC~\cite{fang2024explore} includes two variants: PIC-Cat and PIC-Sep, which differ in how they combine ``input'' and ``target'' point clouds. PIC-Cat concatenates the ``input'' and ``target'' point patches before feeding them into the transformer, while PIC-Sep processes the ``input'' and ``target'' point patches in parallel and merges their features after several blocks. We test our method on both variants.

\subsection{Comparisons with State-of-The-Art Methods}
\begin{figure*}[t]
    \centering
    \includegraphics[width=1.0\linewidth]{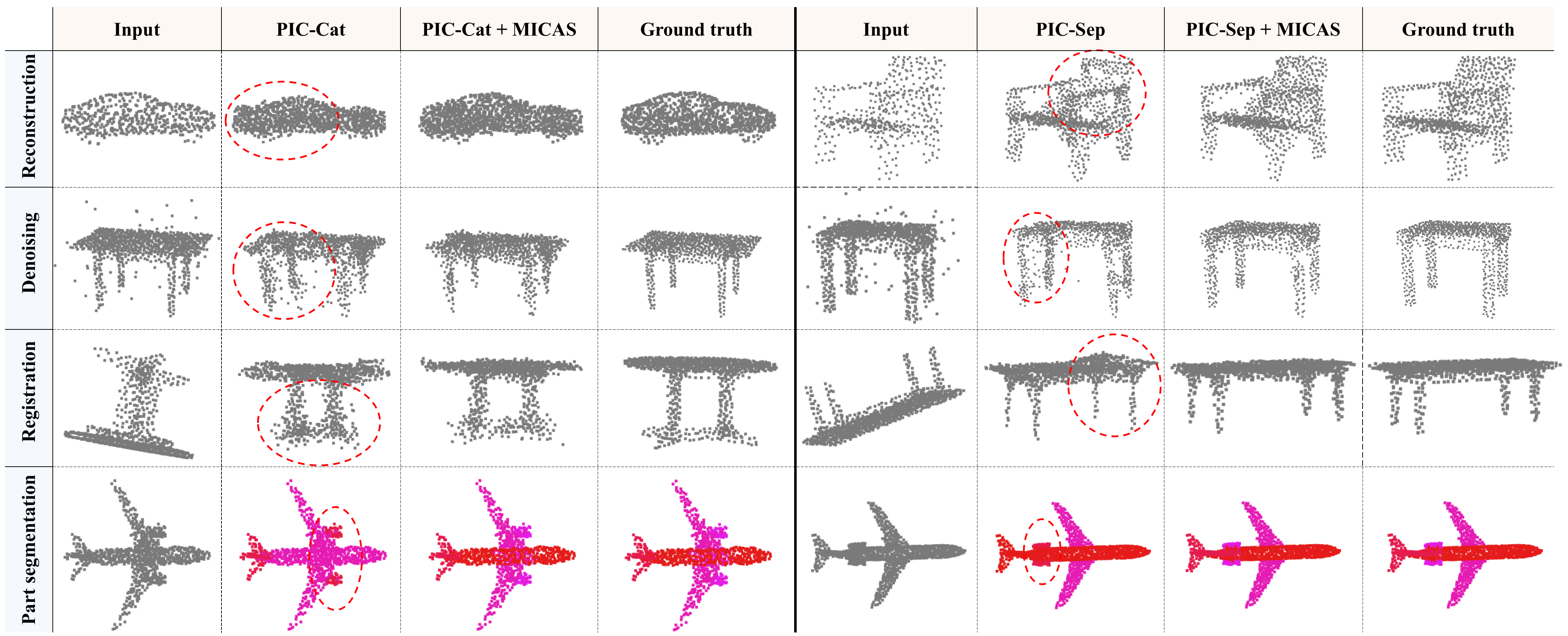}
    \vspace{-1.0em}
    \caption{Qualitative experimental results compared with the PIC-Cat~\cite{fang2024explore} and PIC-Sep~\cite{fang2024explore}. The \red{red ovals} represent the difference between the two methods. Additional visualization results can be found in the supplementary material. (Zoom in for more details)}
    \label{presentation_all}
    \vspace{-1.2em}
  \end{figure*}

We compare our MICAS with various models on the ShapeNet In-Context dataset~\cite{fang2024explore} in Table~\ref{tab:sota} and Figure~\ref{presentation_all}.

\begin{table*}[t]
  \centering
  \caption{Ablation studies on the ShapeNet In-Context Dataset~\cite{fang2024explore}. FPS: farthest point sampling. Point: task-adaptive point sampling. Prompt: query-specific prompt sampling. Inference time represents the average time required to process a query on three 1080ti GPUs.
  }
  \scriptsize
  \setlength\tabcolsep{1.3mm}
  \renewcommand\arraystretch{1.3}
  \begin{tabular}{l|ccc|cccccc|cccccc|cccccc|c|c}
  \toprule[0.15em]
  \multirow{2}{*}{ICL Model} & \multirow{2}{*}{FPS} & \multirow{2}{*}{Point} & \multirow{2}{*}{Prompt} & \multicolumn{6}{c|}{Reconstruction CD $\downarrow$}        & \multicolumn{6}{c|}{Denoising CD $\downarrow$}           & \multicolumn{6}{c|}{Registration CD $\downarrow$}     & Part Seg. & Inference \\
  \multicolumn{1}{l|}{} & \multicolumn{1}{c}{}    & \multicolumn{1}{c}{}    & \multicolumn{1}{c|}{}                   & L1    & L2    & L3    & L4    & L5    & \multicolumn{1}{c|}{Avg.}  & L1    & L2    & L3    & L4    & L5    & \multicolumn{1}{c|}{Avg.}  & L1    & L2    & L3    & L4    & L5    & \multicolumn{1}{c|}{Avg.}  & mIOU$\uparrow$      &    time (ms)    \\ \hline
  \multirow{4}{*}{PIC-Cat~\cite{fang2024explore}} & $\surd$ &  &  & 4.9  &\textbf{\textcolor{blue}{4.1}}   &4.5   & 4.7   & 6.3   &\multicolumn{1}{c|}{4.9}   & 
  \textbf{\textcolor{blue}{4.2}}  & 5.1   & 5.9   & 6.8   & 7.8   & 6.0   & 6.5   & 7.8   & 13.6  & 20.4  & 24.5 & 14.5  &   79.9  & 15.6 \\
  
   & &$\surd$     &  & 4.8 & 4.2 & 4.5 & 4.8 & 5.8 & 4.8 &
  4.3 & 4.5 & 4.7 & \textbf{\textcolor{blue}{4.9}} & 5.2 & 4.7 &
  6.5 & 7.5& 11.1& 16.2& 20.2& 12.3 & 87.6 & 21.4 \\
  
  &$\surd$ &     & $\surd$ & 4.8 & \textbf{\textcolor{blue}{4.1}} & \textbf{\textcolor{blue}{4.4}} & \textbf{\textcolor{blue}{4.6}} & 6.2  & \multicolumn{1}{c|}{4.8} &
  \textbf{\textcolor{blue}{4.2}} & 5.0 & 5.7 & 6.5 & 7.3 & \multicolumn{1}{c|}{5.7} &
  \textbf{\textcolor{blue}{5.5}} & \textbf{\textcolor{blue}{6.5}} & 10.0 & 14.5 & 17.7 & \multicolumn{1}{c|}{10.8} &  80.2 & 44.3 \\
  
  & &$\surd$     & $\surd$ & \textbf{\textcolor{blue}{4.6}} & 4.2 & 4.5 & 4.8 & \textbf{\textcolor{blue}{5.7}} & \multicolumn{1}{c|}{\textbf{\textcolor{blue}{4.7}}} &
  \textbf{\textcolor{blue}{4.2}} & \textbf{\textcolor{blue}{4.4}} & \textbf{\textcolor{blue}{4.6}} & \textbf{\textcolor{blue}{4.9}} & \textbf{\textcolor{blue}{5.1}} & \multicolumn{1}{c|}{\textbf{\textcolor{blue}{4.6}}} &
  5.7 & \textbf{\textcolor{blue}{6.5}} & \textbf{\textcolor{blue}{9.1}} & \textbf{\textcolor{blue}{12.5}} & \textbf{\textcolor{blue}{15.4}} & \multicolumn{1}{c|}{\textbf{\textcolor{blue}{9.8}}} &  \textbf{\textcolor{blue}{87.9}} & 47.1 \\
  \hline
  \multirow{4}{*}{PIC-Sep~\cite{fang2024explore}} & $\surd$ &  &  & 3.9  &3.9   & 3.9   & 4.3   & 6.2   &\multicolumn{1}{c|}{4.4}   
  & 6.2  & 7.2   & 7.7   & 8.2   & 8.3   & 7.5   & 7.6   & 7.8   & 8.4  & 9.0  & 10.0 & 8.6  &   78.7  & 15.0 \\
  
   & &$\surd$     &  & 4.2 & 4.1 & 4.2 & 4.6 & 6.1 & 4.6 &
  4.9 & 5.4 & 5.6 & 6.0 & 6.3 & 5.6 &
  7.6 & 7.4 & 7.8 & 9.2 & 10.7 & 8.5 & 86.6  & 20.9 \\
  
  &$\surd$ &     & $\surd$ & \textbf{\textcolor{blue}{3.6}} & \textbf{\textcolor{blue}{3.7}} & \textbf{\textcolor{blue}{3.8}} & \textbf{\textcolor{blue}{4.1}} & 5.8 & \multicolumn{1}{c|}{\textbf{\textcolor{blue}{4.2}}} &
  5.4 & 6.2 & 6.6 & 7.0 & 7.1 & \multicolumn{1}{c|}{6.5} &
  \textbf{\textcolor{blue}{3.3}} & \textbf{\textcolor{blue}{3.4}} & \textbf{\textcolor{blue}{3.5}} & \textbf{\textcolor{blue}{3.6}} & \textbf{\textcolor{blue}{3.8}} & \multicolumn{1}{c|}{\textbf{\textcolor{blue}{3.5}}} &  79.1 & 44.1 \\ 
  
  & &$\surd$     & $\surd$ & 3.8 & 3.9 & 4.0 & 4.4 & \textbf{\textcolor{blue}{5.6}} & \multicolumn{1}{c|}{4.3} &
  \textbf{\textcolor{blue}{4.4}} & \textbf{\textcolor{blue}{4.9}} & \textbf{\textcolor{blue}{5.2}} & \textbf{\textcolor{blue}{5.5}} & \textbf{\textcolor{blue}{5.7}} & \multicolumn{1}{c|}{\textbf{\textcolor{blue}{5.1}}} &
  3.4 & 3.6 & 3.7 & 3.8 & 4.0 & \multicolumn{1}{c|}{3.7} &  \textbf{\textcolor{blue}{86.8}} & 45.9 \\
  \bottomrule[0.1em]
  \end{tabular}
  \label{tab:ablation}
  \end{table*}

\textbf{Comparison to Task-Specific Models.} As shown in Table~\ref{tab:sota}, task-specific models set a high benchmark, delivering peak performance in reconstruction and denoising tasks due to their specialized design. 
However, these models require a dedicated network for each task, leading to significant complexity and resource demands. 
In contrast, MICAS uses only a prompt to guide a single model across multiple tasks and shows remarkable versatility, particularly excelling in registration and part segmentation.
It outperforms ACT~\cite{dong2022autoencoders} by $2.2$ points in registration and an impressive $6.7$ points in part segmentation, showcasing its effectiveness while offering a more streamlined and efficient solution.

\textbf{Comparison to Multi-task Models.} Our proposed MICAS significantly outperforms state-of-the-art multi-task models across four tasks. Compared to Point-MAE~\cite{pang2022masked}, MICAS achieves better results across all five levels of datasets in the reconstruction, denoising, and registration tasks, thanks to its adaptive sampling mechanisms for task-specific feature extraction. In the part segmentation task, MICAS achieves a remarkable mIOU of $65.3$ higher than I2P-MAE~\cite{zhang2023learning}, demonstrating its effectiveness in handling complex segmentation challenges.

\textbf{Comparison to In-context learning Models.} 
Within the realm of in-context learning for point clouds, two main approaches have emerged: Point-BERT~\cite{yu2022point} and PIC~\cite{fang2024explore}. Therein, PIC includes two variants: $\ast$-Cat and $\ast$-Sep. For $\ast$-Cat methods, although MICAS shows a minor shortfall in the reconstruction compared to PIC-Cat~\cite{pang2022masked}, it significantly outperforms in the denoising, registration, and part segmentation tasks. 
Specifically, MICAS surpasses PIC-Cat~\cite{pang2022masked} by $4.3$ in the registration and $8.9$ in the part segmentation. 
Moreover, MICAS consistently outperforms PIC-S-Cat~\cite{liu2024point} across all evaluation metrics and tasks. 
For $\ast$-Sep methods, MICAS achieves superior performance compared to both PIC-Sep~\cite{pang2022masked} and PIC-S-Sep~\cite{liu2024point} across all metrics and tasks. 
In addition, qualitative results in Figure~\ref{presentation_all} further highlight the effectiveness of our proposed method. 

\subsection{Ablation Study}
To demonstrate the effectiveness of MICAS, we perform an ablation study in Table~\ref{tab:ablation}. The results show that task-adaptive point sampling enhances denoising and part segmentation, while query-specific prompt sampling improves reconstruction and registration. They complement each other in both sampling granularity and overall performance.

\textbf{Task-adaptive Point Sampling}. We replace the farthest point sampling (FPS) used in PIC-Cat \cite{fang2024explore} and PIC-Sep \cite{fang2024explore} with task-adaptive point sampling. While task-adaptive point sampling shows both strengths and limitations compared to FPS in the reconstruction task, it demonstrates clear superiority in the denoising, registration, and part segmentation tasks. Specifically, although task-adaptive point sampling yields an average CD loss that is $0.2$ higher in reconstruction compared to FPS when using PIC-Sep \cite{fang2024explore} as the ICL model, it significantly outperforms FPS across all other metrics and tasks. In addition, our proposed task-adaptive point sampling considerably enhances model performance without noticeably impacting inference speed.

\textbf{Query-specific Prompt Sampling.} We conduct two types of experiments, employing query-specific prompt sampling on FPS and task-adaptive point sampling, respectively. Our experimental results indicate that query-specific prompt sampling enhances overall performance. More importantly, the benefits of query-specific prompt sampling and task-adaptive point sampling are complementary. Specifically, task-adaptive sampling excels in enhancing denoising and part segmentation tasks, while query-specific prompt sampling boosts performance in reconstruction and registration tasks. As shown in Table~\ref{tab:ablation}, combining task-adaptive point sampling with query-specific prompt sampling yields the best overall results, achieving significant performance improvements across all tasks. In addition, we find that these enhancements are achievable with only a threefold increase in inference time.

\section{Conclusion}
We undertake an early effort to address the inter-task and intra-task sensitivity issues arising from lacking context adaptation, spanning both point and prompt levels. Specifically, we propose a Multi-grained In-Context Adaptive Sampling, dubbed MICAS, which includes task-adaptive point sampling and query-specific prompt sampling. 
The former is engineered to interpret task information from diverse prompts and amalgamate it with the original point cloud, enabling a sampling approach that is tailored to each prompt. The latter involves identifying the most relevant prompt for each query, which provides more effective task guidance.
To our knowledge, this represents the inaugural exploration into point cloud sampling within an in-context learning framework at both point and prompt levels.

{
    \small
    \bibliographystyle{ieeenat_fullname}
    \bibliography{main}

\begin{thebibliography}{75}
\providecommand{\natexlab}[1]{#1}
\providecommand{\url}[1]{\texttt{#1}}
\expandafter\ifx\csname urlstyle\endcsname\relax
  \providecommand{\doi}[1]{doi: #1}\else
  \providecommand{\doi}{doi: \begingroup \urlstyle{rm}\Url}\fi

\bibitem[Bar et~al.(2022)Bar, Gandelsman, Darrell, Globerson, and
  Efros]{bar2022visual}
Amir Bar, Yossi Gandelsman, Trevor Darrell, Amir Globerson, and Alexei Efros.
\newblock Visual prompting via image inpainting.
\newblock In \emph{NeurIPS}, 2022.

\bibitem[Bolukbasi et~al.(2017)Bolukbasi, Wang, Dekel, and
  Saligrama]{bolukbasi2017adaptive}
Tolga Bolukbasi, Joseph Wang, Ofer Dekel, and Venkatesh Saligrama.
\newblock Adaptive neural networks for efficient inference.
\newblock In \emph{ICML}, 2017.

\bibitem[Brown et~al.(2020)Brown, Mann, Ryder, Subbiah, Kaplan, Dhariwal,
  Neelakantan, Shyam, Sastry, Askell, et~al.]{brown2020language}
Tom Brown, Benjamin Mann, Nick Ryder, Melanie Subbiah, Jared~D Kaplan, Prafulla
  Dhariwal, Arvind Neelakantan, Pranav Shyam, Girish Sastry, Amanda Askell,
  et~al.
\newblock Language models are few-shot learners.
\newblock In \emph{NeurIPS}, 2020.

\bibitem[Burges(2010)]{burges2010ranknet}
Christopher~JC Burges.
\newblock From ranknet to lambdarank to lambdamart: An overview.
\newblock \emph{Learning}, 2010.

\bibitem[Chang et~al.(2015)Chang, Funkhouser, Guibas, Hanrahan, Huang, Li,
  Savarese, Savva, Song, Su, et~al.]{chang2015shapenet}
Angel~X Chang, Thomas Funkhouser, Leonidas Guibas, Pat Hanrahan, Qixing Huang,
  Zimo Li, Silvio Savarese, Manolis Savva, Shuran Song, Hao Su, et~al.
\newblock Shapenet: An information-rich 3d model repository.
\newblock \emph{arXiv}, 2015.

\bibitem[Chen et~al.(2021)Chen, Luo, Gao, and Hu]{chen2021unsupervised}
Haolan Chen, Shitong Luo, Xiang Gao, and Wei Hu.
\newblock Unsupervised learning of geometric sampling invariant representations
  for 3d point clouds.
\newblock In \emph{ICCV}, 2021.

\bibitem[Cheng et~al.(2021)Cheng, Chen, He, Liu, and Bai]{cheng2021net}
Silin Cheng, Xiwu Chen, Xinwei He, Zhe Liu, and Xiang Bai.
\newblock Pra-net: Point relation-aware network for 3d point cloud analysis.
\newblock \emph{IEEE Transactions on Image Processing}, 30:\penalty0
  4436--4448, 2021.

\bibitem[Dong et~al.(2023)Dong, Qi, Zhang, Zhang, Sun, Ge, Yi, and
  Ma]{dong2022autoencoders}
Runpei Dong, Zekun Qi, Linfeng Zhang, Junbo Zhang, Jianjian Sun, Zheng Ge, Li
  Yi, and Kaisheng Ma.
\newblock Autoencoders as cross-modal teachers: Can pretrained 2d image
  transformers help 3d representation learning?
\newblock In \emph{ICLR}, 2023.

\bibitem[Dovrat et~al.(2019)Dovrat, Lang, and Avidan]{dovrat2019learning}
Oren Dovrat, Itai Lang, and Shai Avidan.
\newblock Learning to sample.
\newblock In \emph{CVPR}, 2019.

\bibitem[Fan et~al.(2017)Fan, Su, and Guibas]{fan2017point}
Haoqiang Fan, Hao Su, and Leonidas~J Guibas.
\newblock A point set generation network for 3d object reconstruction from a
  single image.
\newblock In \emph{CVPR}, 2017.

\bibitem[Fang et~al.(2024)Fang, Li, Li, Buhmann, Loy, and Liu]{fang2024explore}
Zhongbin Fang, Xiangtai Li, Xia Li, Joachim~M Buhmann, Chen~Change Loy, and
  Mengyuan Liu.
\newblock Explore in-context learning for 3d point cloud understanding.
\newblock In \emph{NeurIPS}, 2024.

\bibitem[Fix(1985)]{fix1985discriminatory}
Evelyn Fix.
\newblock \emph{Discriminatory analysis: nonparametric discrimination,
  consistency properties}.
\newblock USAF school of Aviation Medicine, 1985.

\bibitem[Groh et~al.(2018)Groh, Wieschollek, and Lensch]{groh2018flex}
Fabian Groh, Patrick Wieschollek, and Hendrik~PA Lensch.
\newblock Flex-convolution: Million-scale point-cloud learning beyond
  grid-worlds.
\newblock In \emph{ACCV}, 2018.

\bibitem[Guo et~al.(2021)Guo, Cai, Liu, Mu, Martin, and Hu]{guo2021pct}
Meng-Hao Guo, Jun-Xiong Cai, Zheng-Ning Liu, Tai-Jiang Mu, Ralph~R Martin, and
  Shi-Min Hu.
\newblock Pct: Point cloud transformer.
\newblock \emph{Computational Visual Media}, 2021.

\bibitem[Guo et~al.(2023)Guo, Zhang, Zhu, Tang, Ma, Han, Chen, Gao, Li, Li,
  et~al.]{guo2023point}
Ziyu Guo, Renrui Zhang, Xiangyang Zhu, Yiwen Tang, Xianzheng Ma, Jiaming Han,
  Kexin Chen, Peng Gao, Xianzhi Li, Hongsheng Li, et~al.
\newblock Point-bind \& point-llm: Aligning point cloud with multi-modality for
  3d understanding, generation, and instruction following.
\newblock \emph{arXiv preprint arXiv:2309.00615}, 2023.

\bibitem[He et~al.(2025)He, Ding, Jiang, and Wen]{he2025segpoint}
Shuting He, Henghui Ding, Xudong Jiang, and Bihan Wen.
\newblock Segpoint: Segment any point cloud via large language model.
\newblock In \emph{ECCV}, 2025.

\bibitem[Hu et~al.(2020)Hu, Yang, Xie, Rosa, Guo, Wang, Trigoni, and
  Markham]{hu2020randla}
Qingyong Hu, Bo Yang, Linhai Xie, Stefano Rosa, Yulan Guo, Zhihua Wang, Niki
  Trigoni, and Andrew Markham.
\newblock Randla-net: Efficient semantic segmentation of large-scale point
  clouds.
\newblock In \emph{CVPR}, 2020.

\bibitem[Huang et~al.(2024)Huang, Wen, Wang, Ren, and Jia]{huang2024surface}
Zhangjin Huang, Yuxin Wen, Zihao Wang, Jinjuan Ren, and Kui Jia.
\newblock Surface reconstruction from point clouds: A survey and a benchmark.
\newblock \emph{IEEE Transactions on Pattern Analysis and Machine
  Intelligence}, 2024.

\bibitem[Jang et~al.(2016)Jang, Gu, and Poole]{jang2016categorical}
Eric Jang, Shixiang Gu, and Ben Poole.
\newblock Categorical reparameterization with gumbel-softmax.
\newblock In \emph{ICLR}, 2016.

\bibitem[Javaheri et~al.(2017)Javaheri, Brites, Pereira, and
  Ascenso]{javaheri2017subjective}
Alireza Javaheri, Catarina Brites, Fernando Pereira, and Jo{\~a}o Ascenso.
\newblock Subjective and objective quality evaluation of 3d point cloud
  denoising algorithms.
\newblock In \emph{ICMEW}, 2017.

\bibitem[Jenke et~al.(2006)Jenke, Wand, Bokeloh, Schilling, and
  Stra{\ss}er]{jenke2006bayesian}
Philipp Jenke, Michael Wand, Martin Bokeloh, Andreas Schilling, and Wolfgang
  Stra{\ss}er.
\newblock Bayesian point cloud reconstruction.
\newblock In \emph{Computer graphics forum}, 2006.

\bibitem[Jiang et~al.(2025)Jiang, Zhou, Li, Lu, Wang, Ma, Chang, and
  Zhang]{jiang2025dg}
Jincen Jiang, Qianyu Zhou, Yuhang Li, Xuequan Lu, Meili Wang, Lizhuang Ma, Jian
  Chang, and Jian~Jun Zhang.
\newblock Dg-pic: Domain generalized point-in-context learning for point cloud
  understanding.
\newblock In \emph{ECCV}, 2025.

\bibitem[Kim et~al.(2021)Kim, Choe, Yun, and Kwak]{kim2021normalization}
Jeesoo Kim, Junsuk Choe, Sangdoo Yun, and Nojun Kwak.
\newblock Normalization matters in weakly supervised object localization.
\newblock In \emph{ICCV}, 2021.

\bibitem[Kolodiazhnyi et~al.(2024)Kolodiazhnyi, Vorontsova, Konushin, and
  Rukhovich]{kolodiazhnyi2024oneformer3d}
Maxim Kolodiazhnyi, Anna Vorontsova, Anton Konushin, and Danila Rukhovich.
\newblock Oneformer3d: One transformer for unified point cloud segmentation.
\newblock In \emph{CVPR}, 2024.

\bibitem[Landrieu and Simonovsky(2018)]{landrieu2018large}
Loic Landrieu and Martin Simonovsky.
\newblock Large-scale point cloud semantic segmentation with superpoint graphs.
\newblock In \emph{CVPR}, 2018.

\bibitem[Lang et~al.(2020)Lang, Manor, and Avidan]{lang2020samplenet}
Itai Lang, Asaf Manor, and Shai Avidan.
\newblock Samplenet: Differentiable point cloud sampling.
\newblock In \emph{CVPR}, 2020.

\bibitem[Levy et~al.(2023)Levy, Bogin, and Berant]{levy2022diverse}
Itay Levy, Ben Bogin, and Jonathan Berant.
\newblock Diverse demonstrations improve in-context compositional
  generalization.
\newblock In \emph{ACL}, 2023.

\bibitem[Li et~al.(2023)Li, Lv, Yan, Lin, Zhu, Ni, Xie, Wang, and
  Qiu]{li2023unified}
Xiaonan Li, Kai Lv, Hang Yan, Tianyang Lin, Wei Zhu, Yuan Ni, Guotong Xie,
  Xiaoling Wang, and Xipeng Qiu.
\newblock Unified demonstration retriever for in-context learning.
\newblock In \emph{ACL}, 2023.

\bibitem[Li et~al.(2018)Li, Bu, Sun, Wu, Di, and Chen]{li2018pointcnn}
Yangyan Li, Rui Bu, Mingchao Sun, Wei Wu, Xinhan Di, and Baoquan Chen.
\newblock Pointcnn: Convolution on x-transformed points.
\newblock In \emph{NeurIPS}, 2018.

\bibitem[Liu et~al.(2022)Liu, Shen, Zhang, Dolan, Carin, and
  Chen]{liu2021makes}
Jiachang Liu, Dinghan Shen, Yizhe Zhang, Bill Dolan, Lawrence Carin, and Weizhu
  Chen.
\newblock What makes good in-context examples for gpt-$3 $?
\newblock In \emph{DeeLIO}, 2022.

\bibitem[Liu et~al.(2024)Liu, Fang, Li, Buhmann, Li, and Loy]{liu2024point}
Mengyuan Liu, Zhongbin Fang, Xia Li, Joachim~M Buhmann, Xiangtai Li, and
  Chen~Change Loy.
\newblock Point-in-context: Understanding point cloud via in-context learning.
\newblock \emph{arXiv}, 2024.

\bibitem[Liu et~al.(2019)Liu, Fan, Xiang, and Pan]{liu2019relation}
Yongcheng Liu, Bin Fan, Shiming Xiang, and Chunhong Pan.
\newblock Relation-shape convolutional neural network for point cloud analysis.
\newblock In \emph{CVPR}, 2019.

\bibitem[Loshchilov and Hutter(2017)]{loshchilov2016sgdr}
Ilya Loshchilov and Frank Hutter.
\newblock Sgdr: Stochastic gradient descent with warm restarts.
\newblock In \emph{ICLR}, 2017.

\bibitem[Luo and Hu(2021)]{luo2021score}
Shitong Luo and Wei Hu.
\newblock Score-based point cloud denoising.
\newblock In \emph{ICCV}, 2021.

\bibitem[Luo and Yang(2024)]{luo2024large}
Yawei Luo and Yi Yang.
\newblock Large language model and domain-specific model collaboration for
  smart education.
\newblock \emph{Frontiers of Information Technology \& Electronic Engineering},
  25\penalty0 (3):\penalty0 333--341, 2024.

\bibitem[Ma et~al.(2022)Ma, Qin, You, Ran, and Fu]{ma2022rethinking}
Xu Ma, Can Qin, Haoxuan You, Haoxi Ran, and Yun Fu.
\newblock Rethinking network design and local geometry in point cloud: A simple
  residual mlp framework.
\newblock In \emph{ICLR}, 2022.

\bibitem[Mandikal and Radhakrishnan(2019)]{mandikal2019dense}
Priyanka Mandikal and Venkatesh~Babu Radhakrishnan.
\newblock Dense 3d point cloud reconstruction using a deep pyramid network.
\newblock In \emph{WACV}, 2019.

\bibitem[Moryossef et~al.(2019)Moryossef, Goldberg, and
  Dagan]{moryossef2019step}
Amit Moryossef, Yoav Goldberg, and Ido Dagan.
\newblock Step-by-step: Separating planning from realization in neural
  data-to-text generation.
\newblock \emph{arXiv preprint arXiv:1904.03396}, 2019.

\bibitem[Nezhadarya et~al.(2020)Nezhadarya, Taghavi, Razani, Liu, and
  Luo]{nezhadarya2020adaptive}
Ehsan Nezhadarya, Ehsan Taghavi, Ryan Razani, Bingbing Liu, and Jun Luo.
\newblock Adaptive hierarchical down-sampling for point cloud classification.
\newblock In \emph{CVPR}, 2020.

\bibitem[Nichol et~al.(2022)Nichol, Jun, Dhariwal, Mishkin, and
  Chen]{nichol2022point}
Alex Nichol, Heewoo Jun, Prafulla Dhariwal, Pamela Mishkin, and Mark Chen.
\newblock Point-e: A system for generating 3d point clouds from complex
  prompts.
\newblock \emph{arXiv preprint arXiv:2212.08751}, 2022.

\bibitem[Nie et~al.(2023)Nie, Liang, Schmid, and Sch{\"u}tze]{nie2022cross}
Ercong Nie, Sheng Liang, Helmut Schmid, and Hinrich Sch{\"u}tze.
\newblock Cross-lingual retrieval augmented prompt for low-resource languages.
\newblock In \emph{ACL}, 2023.

\bibitem[Pang et~al.(2022)Pang, Wang, Tay, Liu, Tian, and Yuan]{pang2022masked}
Yatian Pang, Wenxiao Wang, Francis~EH Tay, Wei Liu, Yonghong Tian, and Li Yuan.
\newblock Masked autoencoders for point cloud self-supervised learning.
\newblock In \emph{ECCV}, 2022.

\bibitem[Qi et~al.(2017{\natexlab{a}})Qi, Su, Mo, and Guibas]{qi2017pointnet}
Charles~R Qi, Hao Su, Kaichun Mo, and Leonidas~J Guibas.
\newblock Pointnet: Deep learning on point sets for 3d classification and
  segmentation.
\newblock In \emph{CVPR}, 2017{\natexlab{a}}.

\bibitem[Qi et~al.(2017{\natexlab{b}})Qi, Yi, Su, and Guibas]{qi2017pointnet++}
Charles~Ruizhongtai Qi, Li Yi, Hao Su, and Leonidas~J Guibas.
\newblock Pointnet++: Deep hierarchical feature learning on point sets in a
  metric space.
\newblock In \emph{NeurIPS}, 2017{\natexlab{b}}.

\bibitem[Qi et~al.(2023)Qi, Dong, Fan, Ge, Zhang, Ma, and Yi]{qi2023contrast}
Zekun Qi, Runpei Dong, Guofan Fan, Zheng Ge, Xiangyu Zhang, Kaisheng Ma, and Li
  Yi.
\newblock Contrast with reconstruct: Contrastive 3d representation learning
  guided by generative pretraining.
\newblock In \emph{ICML}, 2023.

\bibitem[Radford et~al.(2021)Radford, Kim, Hallacy, Ramesh, Goh, Agarwal,
  Sastry, Askell, Mishkin, Clark, et~al.]{radford2021learning}
Alec Radford, Jong~Wook Kim, Chris Hallacy, Aditya Ramesh, Gabriel Goh,
  Sandhini Agarwal, Girish Sastry, Amanda Askell, Pamela Mishkin, Jack Clark,
  et~al.
\newblock Learning transferable visual models from natural language
  supervision.
\newblock In \emph{ICML}, 2021.

\bibitem[Rubin et~al.(2022)Rubin, Herzig, and Berant]{rubin2021learning}
Ohad Rubin, Jonathan Herzig, and Jonathan Berant.
\newblock Learning to retrieve prompts for in-context learning.
\newblock In \emph{NAACL}, 2022.

\bibitem[Shan et~al.(2023)Shan, Yang, Ye, Zhang, Xu, Xu, and Liu]{shan2023gpa}
Ziyu Shan, Qi Yang, Rui Ye, Yujie Zhang, Yiling Xu, Xiaozhong Xu, and Shan Liu.
\newblock Gpa-net: No-reference point cloud quality assessment with multi-task
  graph convolutional network.
\newblock \emph{TVCG}, 2023.

\bibitem[Shao et~al.(2022)Shao, Luo, Liu, Chen, Yang, Lu, and
  Xiao]{shao2022active}
Feifei Shao, Yawei Luo, Ping Liu, Jie Chen, Yi Yang, Yulei Lu, and Jun Xiao.
\newblock Active learning for point cloud semantic segmentation via
  spatial-structural diversity reasoning.
\newblock In \emph{ACM MM}, 2022.

\bibitem[Tang et~al.(2024)Tang, Han, Li, Yu, Hao, Hu, and
  Chen]{tang2024minigpt}
Yuan Tang, Xu Han, Xianzhi Li, Qiao Yu, Yixue Hao, Long Hu, and Min Chen.
\newblock Minigpt-3d: Efficiently aligning 3d point clouds with large language
  models using 2d priors.
\newblock In \emph{ACM MM}, 2024.

\bibitem[Thomas et~al.(2019)Thomas, Qi, Deschaud, Marcotegui, Goulette, and
  Guibas]{thomas2019kpconv}
Hugues Thomas, Charles~R Qi, Jean-Emmanuel Deschaud, Beatriz Marcotegui,
  Fran{\c{c}}ois Goulette, and Leonidas~J Guibas.
\newblock Kpconv: Flexible and deformable convolution for point clouds.
\newblock In \emph{ICCV}, 2019.

\bibitem[Touvron et~al.(2023)Touvron, Lavril, Izacard, Martinet, Lachaux,
  Lacroix, Rozi{\`e}re, Goyal, Hambro, Azhar, et~al.]{touvron2023llama}
Hugo Touvron, Thibaut Lavril, Gautier Izacard, Xavier Martinet, Marie-Anne
  Lachaux, Timoth{\'e}e Lacroix, Baptiste Rozi{\`e}re, Naman Goyal, Eric
  Hambro, Faisal Azhar, et~al.
\newblock Llama: Open and efficient foundation language models.
\newblock \emph{arXiv preprint arXiv:2302.13971}, 2023.

\bibitem[Vafaei et~al.(2022)Vafaei, Ribeiro, and
  Camarinha-Matos]{vafaei2022assessing}
Nazanin Vafaei, Rita~A Ribeiro, and Luis~M Camarinha-Matos.
\newblock Assessing normalization techniques for simple additive weighting
  method.
\newblock \emph{Procedia Computer Science}, 199:\penalty0 1229--1236, 2022.

\bibitem[Wang et~al.(2023{\natexlab{a}})Wang, Wang, Cao, Shen, and
  Huang]{wang2023images}
Xinlong Wang, Wen Wang, Yue Cao, Chunhua Shen, and Tiejun Huang.
\newblock Images speak in images: A generalist painter for in-context visual
  learning.
\newblock In \emph{CVPR}, 2023{\natexlab{a}}.

\bibitem[Wang et~al.(2023{\natexlab{b}})Wang, Zhang, Cao, Wang, Shen, and
  Huang]{wang2023seggpt}
Xinlong Wang, Xiaosong Zhang, Yue Cao, Wen Wang, Chunhua Shen, and Tiejun
  Huang.
\newblock Seggpt: Segmenting everything in context.
\newblock In \emph{ICCV}, 2023{\natexlab{b}}.

\bibitem[Wang et~al.(2019)Wang, Sun, Liu, Sarma, Bronstein, and
  Solomon]{wang2019dynamic}
Yue Wang, Yongbin Sun, Ziwei Liu, Sanjay~E Sarma, Michael~M Bronstein, and
  Justin~M Solomon.
\newblock Dynamic graph cnn for learning on point clouds.
\newblock \emph{ToG}, 2019.

\bibitem[Wei et~al.(2024)Wei, Chen, Nan, Wang, Qin, and Wei]{wei2024pathnet}
Zeyong Wei, Honghua Chen, Liangliang Nan, Jun Wang, Jing Qin, and Mingqiang
  Wei.
\newblock Pathnet: Path-selective point cloud denoising.
\newblock \emph{IEEE Transactions on Pattern Analysis and Machine
  Intelligence}, 2024.

\bibitem[Wen et~al.(2023)Wen, Yu, and Tao]{wen2023learnable}
Cheng Wen, Baosheng Yu, and Dacheng Tao.
\newblock Learnable skeleton-aware 3d point cloud sampling.
\newblock In \emph{CVPR}, 2023.

\bibitem[Wu et~al.(2023)Wu, Zheng, Pfrommer, and Beyerer]{wu2023attention}
Chengzhi Wu, Junwei Zheng, Julius Pfrommer, and J{\"u}rgen Beyerer.
\newblock Attention-based point cloud edge sampling.
\newblock In \emph{CVPR}, 2023.

\bibitem[Wu et~al.(2021)Wu, Li, Wu, Zhang, and Li]{wu2021indexsample}
Zhenyu Wu, Kun Li, Yuhu Wu, Xin Zhang, and Shengming Li.
\newblock Indexsample: A learnable sampling network in point cloud
  classification.
\newblock In \emph{SICE}, 2021.

\bibitem[Xu et~al.(2025)Xu, Wang, Wang, Chen, Pang, and Lin]{xu2025pointllm}
Runsen Xu, Xiaolong Wang, Tai Wang, Yilun Chen, Jiangmiao Pang, and Dahua Lin.
\newblock Pointllm: Empowering large language models to understand point
  clouds.
\newblock In \emph{ECCV}, 2025.

\bibitem[Xu et~al.(2024)Xu, Liu, Pasupat, Kazemi, et~al.]{xu2024context}
Xin Xu, Yue Liu, Panupong Pasupat, Mehran Kazemi, et~al.
\newblock In-context learning with retrieved demonstrations for language
  models: A survey.
\newblock \emph{arXiv}, 2024.

\bibitem[Yan et~al.(2020)Yan, Zheng, Li, Wang, and Cui]{yan2020pointasnl}
Xu Yan, Chaoda Zheng, Zhen Li, Sheng Wang, and Shuguang Cui.
\newblock Pointasnl: Robust point clouds processing using nonlocal neural
  networks with adaptive sampling.
\newblock In \emph{CVPR}, 2020.

\bibitem[Yang et~al.(2020)Yang, Shi, and Carlone]{yang2020teaser}
Heng Yang, Jingnan Shi, and Luca Carlone.
\newblock Teaser: Fast and certifiable point cloud registration.
\newblock \emph{IEEE Transactions on Robotics}, 2020.

\bibitem[Yang et~al.(2019)Yang, Zhang, Ni, Li, Liu, Zhou, and
  Tian]{yang2019modeling}
Jiancheng Yang, Qiang Zhang, Bingbing Ni, Linguo Li, Jinxian Liu, Mengdie Zhou,
  and Qi Tian.
\newblock Modeling point clouds with self-attention and gumbel subset sampling.
\newblock In \emph{CVPR}, 2019.

\bibitem[Yi et~al.(2016)Yi, Kim, Ceylan, Shen, Yan, Su, Lu, Huang, Sheffer, and
  Guibas]{yi2016scalable}
Li Yi, Vladimir~G Kim, Duygu Ceylan, I-Chao Shen, Mengyan Yan, Hao Su, Cewu Lu,
  Qixing Huang, Alla Sheffer, and Leonidas Guibas.
\newblock A scalable active framework for region annotation in 3d shape
  collections.
\newblock \emph{ToG}, 2016.

\bibitem[Yu et~al.(2023)Yu, Iter, Wang, Xu, Ju, Sanyal, Zhu, Zeng, and
  Jiang]{yu2022generate}
Wenhao Yu, Dan Iter, Shuohang Wang, Yichong Xu, Mingxuan Ju, Soumya Sanyal,
  Chenguang Zhu, Michael Zeng, and Meng Jiang.
\newblock Generate rather than retrieve: Large language models are strong
  context generators.
\newblock In \emph{ICLR}, 2023.

\bibitem[Yu et~al.(2022)Yu, Tang, Rao, Huang, Zhou, and Lu]{yu2022point}
Xumin Yu, Lulu Tang, Yongming Rao, Tiejun Huang, Jie Zhou, and Jiwen Lu.
\newblock Point-bert: Pre-training 3d point cloud transformers with masked
  point modeling.
\newblock In \emph{CVPR}, 2022.

\bibitem[Zeng et~al.(2023)Zeng, Jiang, Mao, Han, Ye, Huang, Yeung, Yang, Liang,
  and Xu]{zeng2023clip2}
Yihan Zeng, Chenhan Jiang, Jiageng Mao, Jianhua Han, Chaoqiang Ye, Qingqiu
  Huang, Dit-Yan Yeung, Zhen Yang, Xiaodan Liang, and Hang Xu.
\newblock Clip2: Contrastive language-image-point pretraining from real-world
  point cloud data.
\newblock In \emph{CVPR}, 2023.

\bibitem[Zhang et~al.(2022)Zhang, Guo, Zhang, Li, Miao, Cui, Qiao, Gao, and
  Li]{zhang2022pointclip}
Renrui Zhang, Ziyu Guo, Wei Zhang, Kunchang Li, Xupeng Miao, Bin Cui, Yu Qiao,
  Peng Gao, and Hongsheng Li.
\newblock Pointclip: Point cloud understanding by clip.
\newblock In \emph{Proceedings of the IEEE/CVF conference on computer vision
  and pattern recognition}, pages 8552--8562, 2022.

\bibitem[Zhang et~al.(2023{\natexlab{a}})Zhang, Wang, Qiao, Gao, and
  Li]{zhang2023learning}
Renrui Zhang, Liuhui Wang, Yu Qiao, Peng Gao, and Hongsheng Li.
\newblock Learning 3d representations from 2d pre-trained models via
  image-to-point masked autoencoders.
\newblock In \emph{CVPR}, 2023{\natexlab{a}}.

\bibitem[Zhang and Yang(2021)]{zhang2021survey}
Yu Zhang and Qiang Yang.
\newblock A survey on multi-task learning.
\newblock \emph{TKDE}, 2021.

\bibitem[Zhang et~al.(2024)Zhang, Wang, Huang, Wang, and Feng]{zhang2024svc}
Yaojie Zhang, Weijun Wang, Tianlun Huang, Zhiyong Wang, and Wei Feng.
\newblock Svc: Sight view constraint for robust point cloud registration.
\newblock \emph{Image and Vision Computing}, page 105315, 2024.

\bibitem[Zhang et~al.(2023{\natexlab{b}})Zhang, Zhang, Li, and
  Smola]{zhang2022automatic}
Zhuosheng Zhang, Aston Zhang, Mu Li, and Alex Smola.
\newblock Automatic chain of thought prompting in large language models.
\newblock In \emph{ICLR}, 2023{\natexlab{b}}.

\bibitem[Zhao et~al.(2024)Zhao, Hu, Yang, Dou, and Kang]{zhao2024robust}
Luda Zhao, Yihua Hu, Xing Yang, Zhenglei Dou, and Linshuang Kang.
\newblock Robust multi-task learning network for complex lidar point cloud data
  preprocessing.
\newblock \emph{Expert Systems with Applications}, 2024.

\end{thebibliography}
}

\clearpage
\setcounter{page}{1}

\setcounter{table}{0}
\renewcommand\thetable{A\arabic{table}}
\setcounter{figure}{0}
\renewcommand{\thefigure}{A\arabic{figure}}  

\maketitlesupplementary

\section*{Summary of the Appendix}
To complement the main paper, this supplementary material provides additional details and insights, structured as follows: 
\begin{itemize}
    \item Sec.~\ref{sec_robustness} presents an additional ablation study to demonstrate the robustness of our proposed MICAS.
    \item Sec.~\ref{sec_sampling} provides further qualitative analysis by showcasing sampled points.
    \item Sec.~\ref{sec_discuss} discusses the limitations of our approach and its broader impacts.
\end{itemize}

\begin{table*}[h]
  \centering
  \caption{Robustness studies on the ShapeNet In-Context Dataset~\cite{fang2024explore}. ICL Model: in-context learning model. FPS: farthest point sampling. Point: task-adaptive point sampling. Prompt: query-specific prompt sampling. Introduced Model: the network model used by the task encoder, point encoder, and prompt sampling module in our proposed MICAS.
  }
  \scriptsize
  \setlength\tabcolsep{1.2mm}
  \renewcommand\arraystretch{1.3}
  \begin{tabular}{l|ccc|cccccc|cccccc|cccccc|c|c}
  \toprule[0.15em]
  \multirow{2}{*}{ICL Model} & \multirow{2}{*}{FPS} & \multirow{2}{*}{Point} & \multirow{2}{*}{Prompt} & \multicolumn{6}{c|}{Reconstruction CD $\downarrow$}        & \multicolumn{6}{c|}{Denoising CD $\downarrow$}           & \multicolumn{6}{c|}{Registration CD $\downarrow$}     & Part Seg. & Introduced \\
  \multicolumn{1}{l|}{} & \multicolumn{1}{c}{}    & \multicolumn{1}{c}{}    & \multicolumn{1}{c|}{}                   & L1    & L2    & L3    & L4    & L5    & \multicolumn{1}{c|}{Avg.}  & L1    & L2    & L3    & L4    & L5    & \multicolumn{1}{c|}{Avg.}  & L1    & L2    & L3    & L4    & L5    & \multicolumn{1}{c|}{Avg.}  & mIOU$\uparrow$      &    Model   \\ \hline
  
  \multirow{6}{*}{PIC-Cat~\cite{fang2024explore}} & $\surd$ &  &  & 4.9  &\textbf{\textcolor{blue}{4.1}}   &\textbf{\textcolor{blue}{4.5}}   & \textbf{\textcolor{blue}{4.7}}   & 6.3   &\multicolumn{1}{c|}{4.9}   & 
  \textbf{\textcolor{blue}{4.2}}  & 5.1   & 5.9   & 6.8   & 7.8   & 6.0   & 6.5   & 7.8   & 13.6  & 20.4  & 24.5 & 14.5  &   79.9  & - \\ 

  & &$\surd$     &  & 4.8 & 4.2 & \textbf{\textcolor{blue}{4.5}} & 4.8 & 5.8 & 4.8 &
  4.3 & 4.5 & 4.7 & \textbf{\textcolor{blue}{4.9}} & 5.2 & 4.7 &
  6.5 & 7.5& 11.1& 16.2& 20.2& 12.3 & 87.6 & PointNet~\cite{qi2017pointnet} \\

  & &$\surd$     & $\surd$ & \textbf{\textcolor{blue}{4.6}} & 4.2 & \textbf{\textcolor{blue}{4.5}} & 4.8 & \textbf{\textcolor{blue}{5.7}} & \multicolumn{1}{c|}{\textbf{\textcolor{blue}{4.7}}} &
  \textbf{\textcolor{blue}{4.2}} & \textbf{\textcolor{blue}{4.4}} & \textbf{\textcolor{blue}{4.6}} & \textbf{\textcolor{blue}{4.9}} & \textbf{\textcolor{blue}{5.1}} & \multicolumn{1}{c|}{\textbf{\textcolor{blue}{4.6}}} &
  \textbf{\textcolor{blue}{5.7}} & \textbf{\textcolor{blue}{6.5}} & \textbf{\textcolor{blue}{9.1}} & \textbf{\textcolor{blue}{12.5}} & \textbf{\textcolor{blue}{15.4}} & \multicolumn{1}{c|}{\textbf{\textcolor{blue}{9.8}}} &  \textbf{\textcolor{blue}{87.9}} & PointNet~\cite{qi2017pointnet} \\ 

    \cline{2-24}
& $\surd$ &  &  & 4.9  &\textbf{\textcolor{blue}{4.1}}   &\textbf{\textcolor{blue}{4.5}}   & \textbf{\textcolor{blue}{4.7}}   & 6.3   &\multicolumn{1}{c|}{\textbf{\textcolor{blue}{4.9}}}   & 
  4.2  & 5.1   & 5.9   & 6.8   & 7.8   & 6.0   & 6.5   & 7.8   & 13.6  & 20.4  & 24.5 & 14.5  &   79.9  & - \\

   & &$\surd$     &  & 4.9 & 4.2 & 4.6 & 4.9 & 5.9 & \textbf{\textcolor{blue}{4.9}} &
  4.1 & 4.3 & 4.6 & \textbf{\textcolor{blue}{4.8}} & 5.0 & 4.6 &
  6.6 & 7.5& 11.5& 16.7& 20.5& 12.6 & \textbf{\textcolor{blue}{85.5}} & DGCNN~\cite{wang2019dynamic} \\

  & &$\surd$     & $\surd$  & 
  \textbf{\textcolor{blue}{4.8}} & 4.2 & 4.6 & 4.9 & \textbf{\textcolor{blue}{5.8}} & \multicolumn{1}{c|}{\textbf{\textcolor{blue}{4.9}}} &
  \textbf{\textcolor{blue}{4.0}} & \textbf{\textcolor{blue}{4.3}} & \textbf{\textcolor{blue}{4.5}} & \textbf{\textcolor{blue}{4.8}} & \textbf{\textcolor{blue}{4.9}} & \multicolumn{1}{c|}{\textbf{\textcolor{blue}{4.5}}} &
  \textbf{\textcolor{blue}{5.8}} & \textbf{\textcolor{blue}{6.7}} & \textbf{\textcolor{blue}{9.5}} & \textbf{\textcolor{blue}{13.0}} & \textbf{\textcolor{blue}{15.9}} & \multicolumn{1}{c|}{\textbf{\textcolor{blue}{10.2}}} &
  85.4 & DGCNN~\cite{wang2019dynamic} \\

  \hline

  \multirow{6}{*}{PIC-Sep~\cite{fang2024explore}}  & $\surd$ &  &  & 3.9  &\textbf{\textcolor{blue}{3.9}}   & \textbf{\textcolor{blue}{3.9}}   & \textbf{\textcolor{blue}{4.3}}   & 6.2   &\multicolumn{1}{c|}{4.4}   
  & 6.2  & 7.2   & 7.7   & 8.2   & 8.3   & 7.5   & 7.6   & 7.8   & 8.4  & 9.0  & 10.0 & 8.6  &   78.7  & -\\ 
  
   & &$\surd$     &  & 4.2 & 4.1 & 4.2 & 4.6 & 6.1 & 4.6 &
  4.9 & 5.4 & 5.6 & 6.0 & 6.3 & 5.6 &
  7.6 & 7.4 & 7.8 & 9.2 & 10.7 & 8.5 & 86.6  & PointNet~\cite{qi2017pointnet} \\

  & &$\surd$     & $\surd$ & \textbf{\textcolor{blue}{3.8}} & \textbf{\textcolor{blue}{3.9}} & 4.0 & 4.4 & \textbf{\textcolor{blue}{5.6}} & \multicolumn{1}{c|}{\textbf{\textcolor{blue}{4.3}}} &
  \textbf{\textcolor{blue}{4.4}} & \textbf{\textcolor{blue}{4.9}} & \textbf{\textcolor{blue}{5.2}} & \textbf{\textcolor{blue}{5.5}} & \textbf{\textcolor{blue}{5.7}} & \multicolumn{1}{c|}{\textbf{\textcolor{blue}{5.1}}} &
  \textbf{\textcolor{blue}{3.4}} & \textbf{\textcolor{blue}{3.6}} & \textbf{\textcolor{blue}{3.7}} & \textbf{\textcolor{blue}{3.8}} & \textbf{\textcolor{blue}{4.0}} & \multicolumn{1}{c|}{\textbf{\textcolor{blue}{3.7}}} &  \textbf{\textcolor{blue}{86.8}} & PointNet~\cite{qi2017pointnet} \\

  \cline{2-24}
   & $\surd$ &  &  & \textbf{\textcolor{blue}{3.9}}  & \textbf{\textcolor{blue}{3.9}}   & \textbf{\textcolor{blue}{3.9}}   & \textbf{\textcolor{blue}{4.3}}   & \textbf{\textcolor{blue}{6.2}}   &\multicolumn{1}{c|}{\textbf{\textcolor{blue}{4.4}}}   
  & 6.2  & 7.2   & 7.7   & 8.2   & 8.3   & 7.5   & 7.6   & 7.8   & 8.4  & 9.0  & 10.0 & 8.6  &   78.7  & -\\ 

   & &$\surd$     &  & 4.4 & 4.2 & 4.3 & 4.9 & 6.7 & 4.9 &
  4.9 & 5.4 & 5.7 & 6.0 & 6.3 & 5.7 &
  8.0 & 8.0& 8.6& 9.3& 9.8& 8.7 & 83.9 & DGCNN~\cite{wang2019dynamic} \\

  & &$\surd$     & $\surd$  & 
  4.0 & 4.0 & 4.2 & 4.6 & \textbf{\textcolor{blue}{6.2}} & \multicolumn{1}{c|}{4.6} &
  \textbf{\textcolor{blue}{4.3}} & \textbf{\textcolor{blue}{4.8}} & \textbf{\textcolor{blue}{5.1}} & \textbf{\textcolor{blue}{5.5}} & \textbf{\textcolor{blue}{5.8}} & \multicolumn{1}{c|}{\textbf{\textcolor{blue}{5.1}}} &
  \textbf{\textcolor{blue}{3.6}} & \textbf{\textcolor{blue}{3.8}} & \textbf{\textcolor{blue}{3.8}} & \textbf{\textcolor{blue}{3.9}} & \textbf{\textcolor{blue}{4.1}} & \multicolumn{1}{c|}{\textbf{\textcolor{blue}{3.9}}} &
  \textbf{\textcolor{blue}{84.0}} & DGCNN~\cite{wang2019dynamic} \\ 
  
  \bottomrule[0.1em]
  \end{tabular}
  \label{tab_dgcnn}
  \end{table*}

\section{Ablation Study: Robustness Analysis}
\label{sec_robustness}

In the task-adaptive point sampling module and query-specific prompt sampling module of our proposed MICAS, we design the task encoder, point encoder, and prompt sampling module based on PointNet~\cite{qi2017pointnet}. To evaluate the robustness of MICAS, we conduct an additional ablation experiment by replacing PointNet with DGCNN~\cite{wang2019dynamic}, a model widely used for CNN-based high-level tasks on point clouds, such as classification and segmentation. Unlike PointNet~\cite{qi2017pointnet}, which relies on a multilayer perceptron (MLP) architecture, DGCNN~\cite{wang2019dynamic} employs a dynamic graph CNN framework and introduces the EdgeConv operation. This operation effectively captures local geometric features of point clouds while maintaining permutation invariance.

The experimental results presented in Table~\ref{tab_dgcnn}, show that the performance trend of MICAS remains consistent across in-context learning models, including PIC-Cat~\cite{fang2024explore} and PIC-Sep~\cite{fang2024explore}, regardless of whether PointNet~\cite{qi2017pointnet} or DGCNN~\cite{wang2019dynamic} is used. These findings highlight the robustness of MICAS, demonstrating its reliability across different in-context learning frameworks and point cloud models.

\section{More Qualitative Analysis }
\label{sec_sampling}

\begin{figure*}
  \centering
  \includegraphics[width=1.0\linewidth]{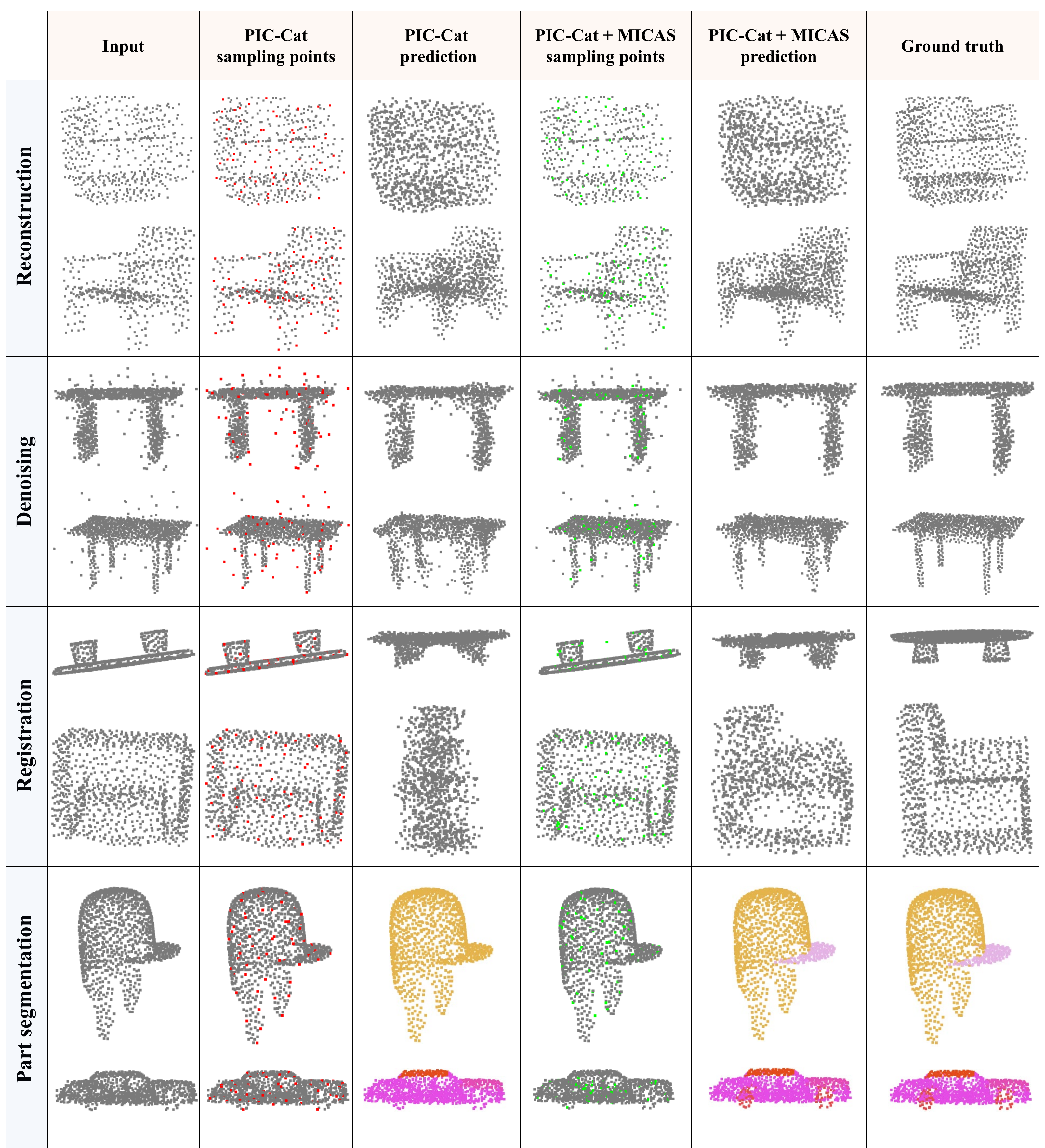}
  \caption{Qualitative experimental results compared with the PIC-Cat~\cite{fang2024explore}. The red and green points denote the central points selected by PIC-Cat and our proposed MICAS, respectively. (Zoom in for more details)}
  \label{presentation_cat}
\end{figure*}

\begin{figure*}
  \centering
  \includegraphics[width=1.0\linewidth]{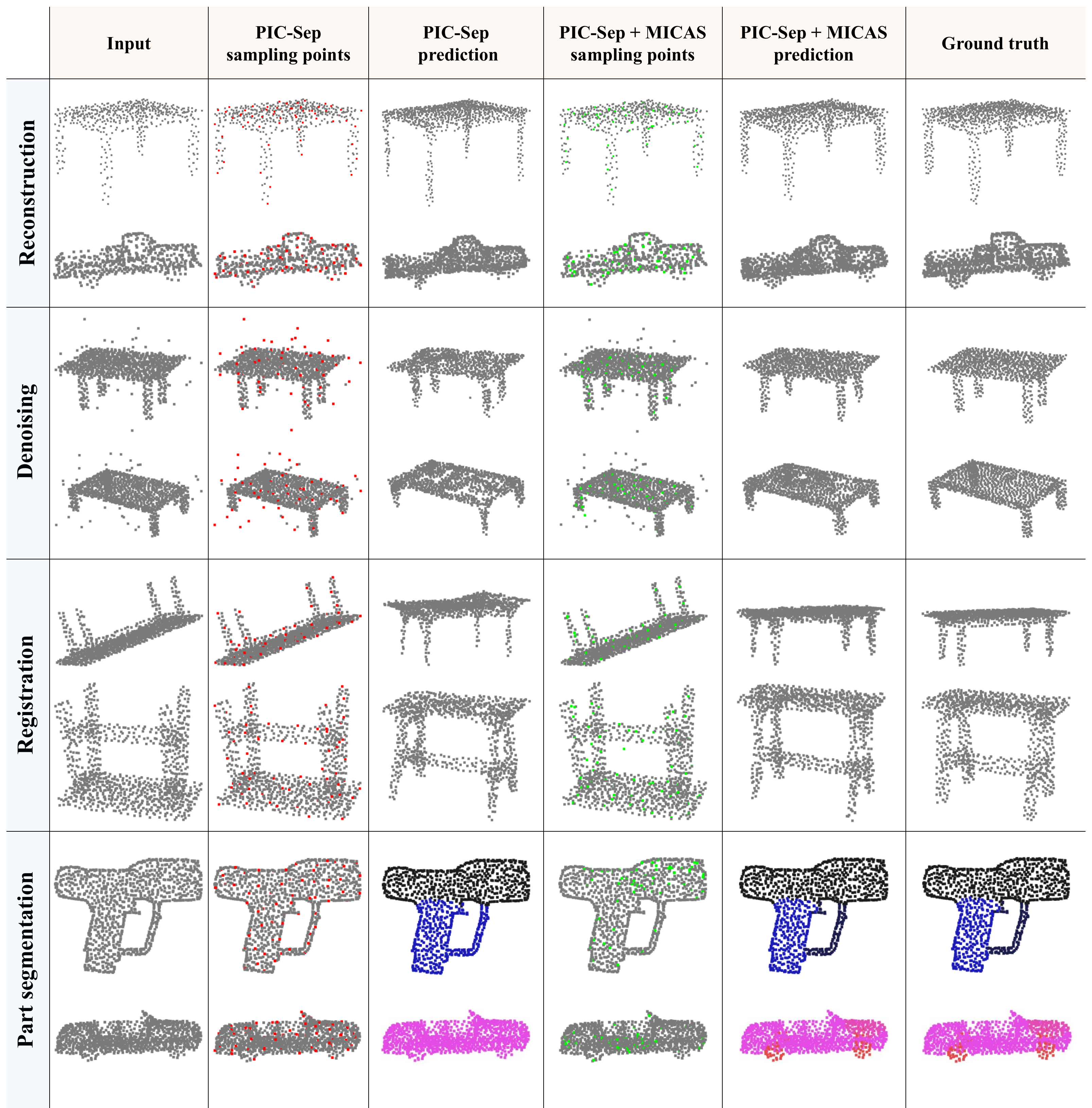}
  \caption{Qualitative experimental results compared with the PIC-Sep~\cite{fang2024explore}. The red and green points denote the central points selected by PIC-Sep and our proposed MICAS, respectively. (Zoom in for more details)}
  \label{presentation_sep}
\end{figure*}

To demonstrate the effectiveness of our proposed MICAS in central point sampling and prediction, we present a visual comparison between our task-adaptive point sampling method and Farthest Point Sampling (FPS) used in PIC-Cat~\cite{fang2024explore} and PIC-Sep~\cite{fang2024explore}. 
As shown in Figures~\ref{presentation_cat} and~\ref{presentation_sep}, our proposed MICAS consistently selects higher-quality central points, delivering superior outcomes and overcoming the limitations of FPS. 
For instance, in the denoising task, FPS often prioritizes outliers, frequently selecting noisy points as central points. In contrast, MICAS effectively avoids these noisy points, focusing on more meaningful and valuable selections. In the reconstruction and registration tasks, MICAS outperforms PIC-Cat~\cite{fang2024explore} and PIC-Sep~\cite{fang2024explore} by producing target point clouds with clearer contours and more accurate shapes. 
Similarly, in the part segmentation task, MICAS achieves accurate segmentation even in areas where PIC-Cat~\cite{fang2024explore} and PIC-Sep~\cite{fang2024explore} encounter segmentation errors.
These visualization results underscore the significance and effectiveness of our proposed MICAS in advancing point cloud in-context learning.

\section{Discussion}
\label{sec_discuss}

\subsection{Limitations}

While our proposed MICAS represents a pioneering effort to address inter-task and intra-task sensitivity challenges in point cloud in-context learning, it has a limitation. Specifically, in the query-specific prompt sampling, we prioritize selecting the ``best-performing'' prompt from a sampled set of $8$ candidate prompts. This process requires predicting the sampling probability for each of the $8$ candidate prompts, which increases the model's inference time. As shown in Table $2$ of the main paper, the query-specific prompt sampling introduces additional computation, adding approximately $25$ ms to the inference time. Nonetheless, despite this slight increase in inference time, the query-specific prompt sampling achieves significant performance gains, particularly in the registration task.

In future work, we recommend addressing this limitation by making the prompt sampling module more lightweight and reducing the size of the prompt candidate pool. Specifically, a simplified prompt sampling module could be developed to streamline the prediction of sampling probabilities and enhance prediction speed. Furthermore, reducing the number of candidate prompts from $8$ to $4$ or even $2$ would significantly lower the computational burden, thereby reducing the overall inference time.

\subsection{Broader Impacts}
This work highlights the limitations of existing learnable task-based sampling approaches~\cite{dovrat2019learning, lang2020samplenet, yan2020pointasnl, wen2023learnable}, which focus solely on inter-point cloud adaptive sampling within the same task and lack the capability to perform inter-task adaptive sampling within the same point cloud. To address this gap, we propose a novel Multi-grained In-Context Adaptive Sampling mechanism, referred to as \textbf{MICAS}, which enables adaptive sampling within the same point cloud by leveraging various prompts.

In summary, our work represents the first shift in point cloud sampling from inter-point cloud adaptive sampling within the same task to inter-task adaptive sampling within the same point cloud. Furthermore, the proposed MICAS contributes positively to the research community by advancing the field of point cloud processing and inspiring future innovations in adaptive in-context learning frameworks.

\end{document}